\documentclass[journal]{IEEEtran}
\usepackage{graphicx}
\usepackage[numbers,sort&compress]{natbib}
\usepackage{amsmath,amssymb,mathtools}
\usepackage{xcolor}
\usepackage{xspace}
\usepackage{xparse}
\usepackage{bbold}
\usepackage{booktabs}
\usepackage{siunitx}
\usepackage{cancel}
\usepackage{enumitem}
\usepackage{csquotes}
\usepackage{dirtytalk}
\usepackage{subfig}
\usepackage{annotate-equations}
\usepackage{hyperref}
\usepackage[capitalize,noabbrev,nameinlink]{cleveref}
\usepackage[acronym,shortcuts]{glossaries}
\usepackage{fontawesome}
\usepackage{float}
\usepackage{cuted}

\makeatletter
\def\tagform@#1{\maketag@@@{\normalsize(#1)}}
\makeatother

\graphicspath{{figures/}}

\newcounter{chapter}

\newcommand{\chapter}[1]{\refstepcounter{chapter}\section{#1}}

\definecolor{intent-orange}{HTML}{ff9500}
\definecolor{intent-blue}{HTML}{0048a6}
\definecolor{intent-teal}{HTML}{0d948f}
\definecolor{intent-purple}{HTML}{800080}
\definecolor{intent-skyblue}{HTML}{0091ff}

\newcommand{\para}[1]{\medskip\noindent\textbf{#1. }}
\newcommand{\inlinecolorbox}[2]{%
  \begingroup
  \setlength{\fboxsep}{0pt}%
  \colorbox{#1}{{\strut #2\strut}}%
  \endgroup
}

\glsdisablehyper

\newacronym{codi}{CoDi}{Coordinated Diffusion}
\newacronym{sde}{SDE}{stochastic differential equation}
\newacronym{dsm}{DSM}{denoising score matching}
\newacronym{il}{IL}{imitation learning}
\newacronym{kl}{KL}{Kullback--Leibler}
\newacronym{ctde}{CTDE}{centralized training with decentralized execution}
\newacronym{dpmd}{DPMD}{Diffusion Policy Mirror Descent}
\newacronym{sdac}{SDAC}{Soft Diffusion Actor-Critic}
\newacronym{expo}{EXPO}{Expressive Policy Optimization}
\newacronym{mppi}{MPPI}{Model Predictive Path Integral Control}
\newacronym{gan}{GAN}{generative adversarial network}

\newcommand{\defeq}{\coloneqq}
\newcommand{\norm}[1]{\left\lVert#1\right\rVert}
\newcommand{\given}{\mid}
\newcommand{\identity}{\mathbf{I}}
\newcommand{\transpose}{\top}
\newcommand{\reals}{\mathbb{R}}
\DeclareMathOperator*{\expectation}{\mathbb{E}}
\DeclareMathOperator*{\covariance}{\mathrm{Cov}}

\newcommand{\cost}{J}
\newcommand{\weight}{w}
\newcommand{\nmppisamples}{M}

\newcommand{\ptarget}{{p}^*}
\newcommand{\sigmadistribution}{\mathcal{W}}
\newcommand{\dataset}{\mathcal{D}}
\newcommand{\madataset}{\mathcal{D}_\text{MA}}
\newcommand{\sample}{\mathbf{x}}
\newcommand{\samplespace}{\mathcal{X}}
\newcommand{\guidecontext}{\mathbf{y}}
\newcommand{\noise}{\mathbf{\epsilon}}
\newcommand{\normal}{\mathcal{N}}
\newcommand{\intd}{\mathrm{d}}
\newcommand{\kldiv}[2]{D_{\text{KL}}\left(#1 \middle\| #2\right)}
\newcommand{\score}{s}
\newcommand{\scoreparams}{\theta}
\NewDocumentCommand{\pdiffusion}{o}{p_{\scoreparams\IfValueT{#1}{, #1}}}
\newcommand{\truescore}{\nabla_{\sample}\log\ptarget_{t}}
\newcommand{\perturbationkernelsym}{q}
\NewDocumentCommand{\scoremodel}{O{} O{\scoreparams}}{
  \score_{#2}
  \IfValueT{#1}{^{#1}}
}
\NewDocumentCommand{\perturbationkernel}{O{t} o m}{
\perturbationkernelsym_{#1}(#3(#1)\given\IfValueTF{#2}{#2}{#3})
}
\newcommand{\productscoremodel}{\scoremodel[(1:N)]}
\newcommand{\wiener}{{\mathbf{w}}}
\newcommand{\wienerback}{\bar\wiener}
\newcommand{\loss}{\mathcal{L}}
\newcommand{\sampledim}{{n_x}}
\newcommand{\guidecontextdim}{{n_y}}
\newcommand{\statedim}{{n_s}}
\newcommand{\actiondim}{{n_a}}
\newcommand{\horizon}{K}
\newcommand{\diffusionprior}{\normal(\sample; \mathbf{0}, T^2\identity)}
\newcommand{\guideparams}{\psi}
\newcommand{\classifier}{p^{\mathrm{c}}}

\newcommand{\policy}{\pi}
\newcommand{\matargetpolicy}{{\policy^*}}
\newcommand{\mapolicy}{{\policy}}
\newcommand{\action}{\mathbf{a}}
\newcommand{\state}{\mathbf{s}}
\newcommand{\actionspace}{\mathcal{A}}
\newcommand{\statespace}{\mathcal{S}}
\newcommand{\jointaction}{\action}
\newcommand{\jointstate}{\state}
\newcommand{\jointactionspace}{\actionspace}
\newcommand{\jointstatespace}{\statespace}
\newcommand{\madiffusionpolicy}{\policy}
\newcommand{\distributionspace}{\mathcal{P}}
\NewDocumentCommand{\mapolicyspace}{o}{\distributionspace\{
  \IfValueTF{#1}{\actionspace^{(#1)}}{\jointactionspace}
  \given\jointstatespace\}}
\newcommand{\mascoremodel}[1][\scoreparams]{\scoremodel[\mapolicy][#1]}
\newcommand{\oursnormalizingconstant}{Z}
\newcommand{\oursregularizationfactor}{\lambda}
\newcommand{\guidancecompensation}{\gamma}

\newcommand{\guide}{J}
\newcommand{\guidancescore}{g^\guide}
\newcommand{\onlineguidancescoreestimate}{\guidancescore_\text{codi}}
\newcommand{\muonlineguidance}{{\mu_\text{codi}}}
\newcommand{\sigmaonlineguidance}{{\Sigma_\text{codi}}}
\NewDocumentCommand{\madatadriven}{o}{\IfValueTF{#1}{\pdiffusion[#1]}{\pdiffusion}^{\scalebox{0.8}{$\scriptstyle 1\text{:}N$}}}
\newcommand{\statedecomposer}[1]{f_\textrm{single}^{(#1)}}
\newcommand{\trajectory}{\tau}
\newcommand{\jointtrajectory}{\trajectory}
\newcommand{\montecarlosamplesize}{M}

\newcommand{\costweight}[1][]{w_\text{#1}}
\newcommand{\costcomponent}[1][]{J_\text{#1}}
\newcommand{\weightedcostcomponent}[1][]{\costweight[#1]\costcomponent[#1]}

\newcommand{\pdist}{d^\text{ws}}
\newcommand{\tarjToPointDistance}{d_{\text{t2p}}}
\newcommand{\trajToTrajDistance}{d_{\text{t2t}}}
\newcommand{\allowedminimumdistance}{\Delta_\text{collision}}
\newcommand{\engageradius}{\Delta_\text{engage}}
\newcommand{\goalradius}{\Delta_\text{goal}}

\newcommand{\costmodelparameters}{\phi}
\newcommand{\costmodel}{J_\costmodelparameters}
\newcommand{\costmodelloss}{\mathcal{L}_\text{CG}}

\begin{document}

\title{Coordinated Diffusion: Generating Multi-Agent Behavior Without Multi-Agent Demonstrations}

\author{%
  \IEEEauthorblockN{Lasse Peters\IEEEauthorrefmark{1},
                    Laura Ferranti\IEEEauthorrefmark{2}\IEEEauthorrefmark{4},
                    Andrea Bajcsy\IEEEauthorrefmark{3}\IEEEauthorrefmark{4},
                    Javier Alonso-Mora\IEEEauthorrefmark{2}\IEEEauthorrefmark{4}}\\
  \IEEEauthorblockA{\IEEEauthorrefmark{1}University of California, Berkeley}
  \IEEEauthorblockA{\IEEEauthorrefmark{2}Delft University of Technology}
  \IEEEauthorblockA{\IEEEauthorrefmark{3}Carnegie Mellon University}
  \thanks{\IEEEauthorrefmark{4}Equal advising.}
}

\maketitle

\begin{strip}
  \vspace{-4em}
  \centering
  \includegraphics[width=\textwidth]{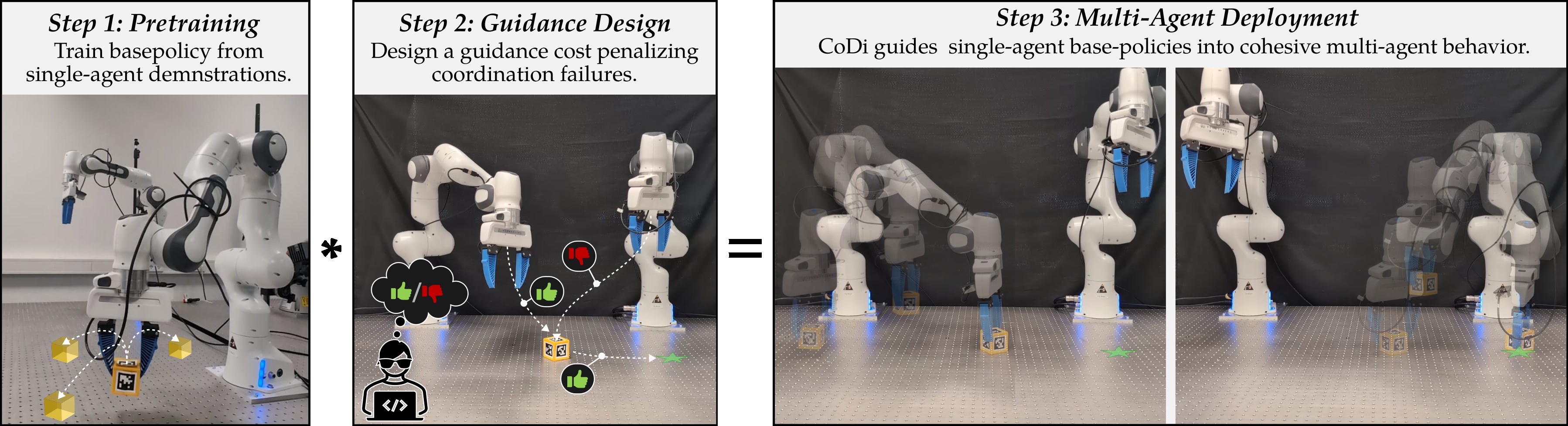}
  \captionof{figure}{\emph{\acf{codi}} is a method for generating multi-agent behavior from single-agent demonstrations.
  \vspace{-1em}
  }
  \label{fig:method-overview}
\end{strip}
\begin{abstract}
  Imitation learning powered by generative models has proven effective for modeling complex single-agent behaviors.
  However, teaching multi-agent systems, like multiple arms or vehicles, to coordinate through imitation learning is hindered by a fundamental data bottleneck: as the joint state-action space grows exponentially with the number of agents, collecting a sufficient amount of coordinated multi-agent demonstrations becomes extremely costly.
  In this work, we ask: how can we leverage \textit{single-agent} demonstration data to learn \textit{multi-agent} policies?
  We present \ac{codi}, a framework that couples independently trained single-agent diffusion policies through a user-defined multi-agent cost function, without requiring any coordinated demonstrations.
  We derive a new diffusion-based sampling scheme wherein the diffusion score function decomposes into independent, single-agent pre-trained base policies plus a cost-driven guidance term that coordinates these base policies into cohesive multi-agent behavior.
  We show that this guidance term can be estimated in a gradient-free manner, making \ac{codi} applicable to black-box, non-differentiable cost functions without additional training.
  Theoretically and empirically, we analyze the conditions under which this composition can faithfully approximate a target multi-agent behavior.
  We find a complementary role for demonstration data versus the cost function: single-agent demonstrations must cover the support of the desired multi-agent behavior, while the cost function must promote desired behavior from this product of single-agent policies.
  Our results in simulation and hardware experiments of a two-arm manipulation task show that \ac{codi} discovers robust coordinated behavior from single-agent data, is more data-efficient than multi-agent baselines, and highlights the importance of joint guidance, base-policy support, and cost design.\\[1em]
  Code: \url{https://autonomousrobots.nl/paper_websites/codi}.
\end{abstract}


\section{Introduction}\label{sec:introduction}

\Ac{il}, combined with advances in generative modeling such as diffusion policies ~\cite{chi2023diffusion-rss,chi2025diffusion-ijrr,janner2022planning,reuss2023goal}, has become a powerful way for single-agent robotic systems to learn complex behaviors from expert demonstrations.
However, many real-world tasks such as tabletop cleaning or furniture assembly require multi-agent interaction: for example, consider the two fixed-base robotic manipulators in \cref{fig:method-overview} where each arm must rely on the other to grab an out-of-reach object.

A key challenge in solving complex multi-agent tasks via imitation is data scarcity, which can be attributed to two main issues: (\romannumeral1) the product space of states and actions grows exponentially with the number of agents, making it challenging to get sufficient coverage and (\romannumeral2) the logistical challenges of acquiring data for multi-agent systems are inevitably higher than for single-agent systems (e.g., simultaneously controlling 2 or more robotic arms).
Meanwhile, the availability of single-agent expert demonstrations has been rapidly increasing via large-scale datasets like Open-X Embodiment~\cite{o2024open} and DROID~\cite{khazatsky2024droid}. 
To tackle these challenges, in this work we ask:
\begin{quote}
  \centering
  \textit{How can we leverage single-agent demonstrations to learn multi-agent policies?}
\end{quote}
Our key insight is that multi-agent success largely relies on low-level capabilities also found in single-agent tasks; the added challenge stems from high-level \emph{coordination} of these single-agent skills.
In the out-of-reach pick-and-place setup, for instance, the core challenge lies in the coordination between agents (e.g., which robot should pick up the object and where to place it next) rather than on the low-level skill of object interaction (e.g., \emph{how} to pick an object).
As a result, single-agent data can serve as a strong prior for learning multi-agent policies.\footnote{
  While we will use multi-agent manipulation as a running example throughout this paper, the method applies more generally to multi-agent tasks in other domains, like navigation.
}

In this paper we introduce \acf{codi}, a novel method for generating samples from the multi-agent behavior distribution by \emph{guiding} multiple single-agent diffusion policies towards \emph{coordinated} behavior.
\ac{codi} follows a three-stage procedure summarized in \cref{fig:method-overview}.
First, as in standard imitation learning, a user provides single-agent demonstrations of basic single-agent skills (e.g., picking and placing objects). This data is used to train $N$ single-agent diffusion policies (base policies) that model the low-level skills of each agent.
Next, the user designs a multi-agent cost function that incentivizes desired multi-agent behavior (e.g., rewarding object movement towards the goal location while penalizing collisions with other agents).
We construct a KL-regularized \emph{cooperative game} from these ingredients, in which agents jointly minimize their expected cost while regularizing towards their single-agent base policies.
We show that this game's solution can bridge the gap between the marginal (single-agent) diffusion policies and desired multi-agent behavior, and propose a centralized guidance model to sample from it at deployment time.

By construction, \ac{codi} effectively inverts the conventional \emph{centralized training, decentralized execution} paradigm~\cite{oliehoek2008optimal}: we enable \emph{decentralized} (single-agent) pre-training at the expense of requiring \emph{centralized} execution.
We contend that in many settings, this trade-off is preferable; enabling centralized execution via multi-agent communication is often less demanding than collecting multi-agent demonstration data.

In simulation and hardware experiments with a collaborative two-arm manipulation task, we validate the following key claims: \textbf{(C1)}~\ac{codi} discovers collaboration strategies from single-agent demonstrations, \textbf{(C2)}~is robust enough for hardware deployment, and \textbf{(C3)}~is more data-efficient than methods trained on multi-agent demonstrations; \textbf{(C4)}~joint guidance is critical for safe and efficient coordination; \textbf{(C5)}~our proposed guidance estimator is necessary for effective coordination; and \textbf{(C6)}~guided behavior depends on the base-policy support, making cost design most critical when coordination is absent from the pre-training data.

\section{Related Work}\label{sec:related_work}

\para{Diffusion Generative Models}
A diffusion model is a neural-network-parameterized stochastic process trained to reverse the corruption of data with noise.
This enables sampling from the complex data distribution by starting from a simple (Gaussian) noise distribution ~\cite{sohl2015deep,ho2020denoising,song2019generative,song2020score}.
While early work formulated the diffusion generative process in discrete time~\cite{sohl2015deep}, more recent approaches formulate denoising dynamics in continuous time~\cite{song2019generative,song2020score,karras2022elucidating}.
\citet{albergo2025stochastic} and \citet{gao2025diffusionmeetsflow} show that the subsequently developed flow-matching framework~\cite{lipman2023flow} can be understood as an alternative perspective on the same underlying paradigm.
Originally popularized in image generation, where diffusion models quickly outperformed \acp{gan}~\cite{dhariwal2021diffusion,karras2022elucidating}, these models have since been extended to many other domains, including audio~\cite{kong2020diffwave}, video~\cite{ho2022video}, and 3D point clouds~\cite{luo2021diffusion}.
In this work, we adopt a continuous-time formulation of diffusion that closely follows the score-based perspectives of \citet{song2020score} and \citet{karras2022elucidating}, apply these models to multi-agent policy learning, a problem that we review next.

\para{Diffusion Policies for Imitation Learning}
Diffusion models have been widely adopted for robotic policy learning, modeling action and state-trajectory distributions to capture expressive, multimodal behavior in visuomotor control, planning, and goal-conditioned imitation~\cite{chi2023diffusion-rss,chi2025diffusion-ijrr,janner2022planning,reuss2023goal}.
In imitation-learning settings, these policies are trained with the same score-matching objective used in image generation.
They have also been extended to multi-agent settings to learn the \emph{joint} distribution over multi-agent trajectories, enabling coordination in tasks such as multi-agent manipulation and multi-robot navigation~\cite{dong2025mimicd,he2025latenttom}.
These priors on multi-agent diffusion rely on \emph{multi-agent} demonstrations and achieve decentralization at test-time via the \ac{ctde} paradigm~\cite{oliehoek2008optimal}.
Our work falls within the multi-agent policy learning setting but inverts the \ac{ctde} training paradigm: we pre-train on single-agent rather than multi-agent demonstrations (facilitating independent training), and use a cost function to centrally coordinate these single-agent policies at test-time into cohesive multi-agent behavior.

\para{Factored Diffusion Models}
Prior work has investigated approximating complex target distributions as products of simpler diffusion-based factors, for example in image generation~\cite{liu2022composable,du2023reduce}.
In single-agent policy learning, this idea has been applied to compose complex skills from simpler base policies~\cite{wang2024poco}.
Our work applies this factorization approach along a different axis: we decompose the joint \emph{multi-agent} distribution into a product of \emph{single-agent} marginals, enabling training from single-agent rather than multi-agent demonstrations.
In this respect, the closest prior work is \citet{shaoul2025multi}, which also leverages single-agent diffusion policies for multi-agent motion planning.
However, their method constructs joint motion plans through \emph{sequential} conditioning and is limited to collision-avoidance settings because it fundamentally requires that each state component be influenced by only one robot's action.
\Ac{codi}, by contrast, guides single-agent policies \emph{simultaneously}, allowing it to be applied to settings in which multiple robots can influence the same state component, as in the manipulation tasks we study.

\para{Diffusion Beyond Imitation: Learning from Reward Feedback}
Imitation learning via a denoising score-matching loss is not the only way to train a diffusion policy.
More recent works go beyond imitation objectives, instead learning diffusion policy parameters from reward signals~\cite{ma2025efficient,dong2025expo}; either from scratch~\cite{ma2025efficient}, or by fine-tuning a pre-trained imitation policy~\cite{dong2025expo,ren2025diffusion}.
Extensions of this diffusion model training paradigm to a multi-agent \ac{ctde} paradigm have also been explored~\cite{li2026diffusecoord,zhu2024madiff}.
In our work, we pre-train on single-agent demonstrations via a vanilla denoising score-matching objective and use reward feedback for guidance (reviewed below), rather than fine-tuning.

\para{Guidance for Diffusion}
Diffusion models admit test-time steering by modifying the denoising dynamics, enabling conditioning on additional information without retraining~\cite{dhariwal2021diffusion,ho2022classifierfree}.
In image generation, this property is widely exploited to improve sample quality and to condition on class labels or text prompts~\cite{dhariwal2021diffusion,ho2022classifierfree,ramesh2022hierarchical}.
Most relevant to our work is classifier guidance~\cite{dhariwal2021diffusion}, which steers generation via gradients of a learned classifier.
In single-agent policy learning, analogous techniques have been applied to guide policies via reward signals~\cite{janner2022planning}.
In our work, we develop a guidance scheme closely related to classifier guidance, with two key differences.
First, we forgo training a classifier altogether: instead of computing guidance via the gradient of a learned classifier through automatic differentiation, we approximate the guidance signal directly from cost evaluations of online samples in a gradient-free manner.
Second, rather than guiding a single monolithic policy, we apply guidance to a \emph{product} of single-agent diffusion policies, enabling coordination without requiring multi-agent demonstrations.

\section{
  Background: Forward- and Reverse-time Diffusion Processes
 }\label{sec:background}
This section provides a brief background on diffusion policies and their theoretical underpinnings.
For a comprehensive treatment of these foundations, we refer the reader to the tutorial by \citet{lai2025principles}.

\subsection{Forward-time Diffusion Processes: From Data to Noise}
\label{sec:background-forward-diffusion}
Forward-time diffusion processes model the act of incrementally corrupting \emph{structured data}~$\sample\in\samplespace\subseteq\reals^\sampledim$ drawn from a target distribution $\ptarget(\sample)$ with \emph{noise}.
To formalize this, let $t \in [0,T]$ denote a virtual \emph{noise-time} that increases as noise is added (so that $t=0$ is associated with noise-free data), and let $\perturbationkernel{\sample}$ denote a time-varying perturbation kernel.
Given these quantities, noise-corrupted data at time~$t$ satisfies:
\begin{align}\label{eq:forward-solution}
  \sample(t) & \sim \ptarget_{t}(\sample(t)) \defeq \int_\samplespace \ptarget(\sample)\perturbationkernel{\sample} \intd\sample.
\end{align}
When choosing a \emph{Gaussian} perturbation kernel, i.e.,
$\perturbationkernel{\sample} \defeq \normal(\sample(t); \sample, t^2\identity)$,
at monotonically increasing isotropic variance $t^2$, 
the process~$\{\sample(t)\}_{t\in[0,T]}$ satisfies the \ac{sde}
\begin{align}
  \label{eq:forward-sde}
  \intd \sample(t)
  = \sqrt{2t} \intd \wiener_t,
\end{align}
where $\wiener_t$ is a standard Wiener process, and the initial condition of the process, $\sample(0)$, follows the target distribution since we have $\ptarget\equiv\ptarget_{0}$.

\subsection{Reverse-time Diffusion Processes: From Noise to Data}
\label{sec:background-reverse-diffusion}
Generative diffusion models~\cite{sohl2015deep,song2019generative,song2020score,ho2020denoising} are based on the idea of \emph{reversing} the time direction in~\cref{eq:forward-sde}.
Intuitively, while the forward-time diffusion process takes structured data as input and adds noise to it (as in \cref{eq:forward-sde}) until it is indistinguishable from a (high-variance) Gaussian, the reverse-time diffusion process takes \emph{noise} as input and transforms it into samples from the structured data distribution $\ptarget$.

This time reversal is enabled by the result of~\citet{anderson1982reverse}, which establishes that the following reverse-time \ac{sde} governs the noise-corrupted data $\sample(t)$ as noise-time flows backwards:
\begin{align}
  \label{eq:reverse-sde}
  \intd \sample(t) & = -2t\truescore(\sample(t)) \intd t + \sqrt{2t} \intd \wienerback_t,\nonumber \\
  \sample(T)       & \sim \ptarget_{T}(\sample(T)).
\end{align}
Here, $\wienerback_t$ is a standard Wiener process when time flows backwards and
$\intd t$ is a \emph{negative} time differential.
Generative diffusion models leverage this \ac{sde} to sample from the target distribution $\ptarget(\sample)$ by learning a model~$\scoremodel(\sample(t); t)$ of the only unknown in this equation: the term $\truescore(\sample(t))$---commonly referred to as the \emph{score}.

\para{Training}
\citet{vincent2011connection} show that, when samples from the data distribution $\ptarget(\sample)$ are available, a \emph{score model}~$\scoremodel(\sample(t); t)$ can be trained by minimizing the \ac{dsm} loss
\begin{align}
  \label{eq:score-matching-loss-simplified}
   \loss(\scoreparams) \defeq \expectation_{\substack{
    \sample \sim \ptarget(\sample)\\
    t \sim \sigmadistribution\\
    \noise \sim \normal(\mathbf{0}, \identity)}}
  \left[
  \norm{
  t^2\scoremodel(\sample + t\noise; t) + t\noise
  }^2
  \right]
\end{align}
where $\scoreparams$ are the trainable parameters of the score model~$\scoremodel$, and $\sigmadistribution$ is a distribution over the noise-time~$t$ following~\cite{karras2022elucidating}.

\para{Generating Samples}
Once the score model is trained, it satisfies $\scoremodel(\sample(t); t) \approx \truescore(\sample(t))$.
Additionally, when the terminal noise scale $T$ is sufficiently large, we have $\ptarget_{T}(\sample(T)) \approx \normal(\sample(T); \mathbf{0}, T^2\identity)$.
Therefore, we can generate samples from the target distribution $\ptarget(\sample)$ by integrating the reverse-time \ac{sde}
\begin{align}
  \label{eq:reverse-sde-with-score-model}
  \intd \sample(t) & = -2t\overbrace{\scoremodel(\sample(t); t)}^{\approx \truescore(\sample(t))} \intd t + \sqrt{2t} \intd \wienerback_t,\nonumber \\
  \sample(T)       & \sim \underbrace{\diffusionprior}_{\approx \ptarget_{T}(\sample(T))}
\end{align}
backwards in time from $T$ to $0$ (cf. \cref{eq:reverse-sde}).

\para{Remarks on Notation and Properties}
\Cref{eq:reverse-sde} defines an implicit distribution over the target domain $\samplespace$ that depends on the score model $\scoremodel$.
We denote this implicit distribution $\pdiffusion(\sample)$.
Note that, while we can draw samples $\sample\sim\pdiffusion(\sample)$ via~\cref{eq:reverse-sde-with-score-model}, we cannot evaluate its density $\pdiffusion(\cdot)$ analytically.

\subsection{Diffusion Policies for Robotics}
\label{sec:background-diffusion-policies}
While diffusion models have been popularized in the context of image generation~\cite{ho2020denoising,dhariwal2021diffusion}, in this work we will use them to represent a stochastic policy for decision-making as in~\cite{janner2022planning,chi2023diffusion-rss,reuss2023goal,chi2025diffusion-ijrr}; i.e., a distribution over \emph{action sequences} (chunks) $\action\in\actionspace\subseteq\reals^{\actiondim\times\horizon}$ conditioned on the state $\state\in\statespace\subseteq\reals^\statedim$, $\pdiffusion(\action\given\state)$.

One simple way of training \emph{diffusion policies} is via \ac{il}: given a dataset $\dataset\defeq\{(\state_{(d)}, \action_{(d)})\}_{d=1}^D$ of $D$ state-action pairs $(\state_{(d)}, \action_{(d)})$ with $\action_{(d)}\sim\policy^*(\action\given\state_{(d)})$, one can train a score model $\scoremodel(\action(t); t, \state)$ via \ac{dsm} as in \cref{eq:score-matching-loss-simplified} whose implicit distribution $\pdiffusion(\action\given\state)$ approximates the true distribution $\policy^*(\action\given\state)$.
Mirroring the sampling procedure of~\cref{sec:background-reverse-diffusion}, sampling actions from the policy amounts to integrating the reverse-time \ac{sde} with the learned score model for a given state~$\state$, which approximates the true distribution~$\policy^*$.

\section{Composing Multi-Agent Behavior from Single-Agent Diffusion Policies}\label{sec:method}

In this work, we seek to synthesize policies that allow \emph{multiple} agents to \emph{collaboratively} interact with each other.
In theory, the concept of diffusion policies introduced in \cref{sec:background-diffusion-policies} naturally extends to such a multi-agent setting.
However, naive training of such a multi-agent diffusion policy via \ac{il} would require vast amounts of \emph{multi-agent} demonstrations: expert demonstrations of how \emph{all} agents should behave \emph{jointly} in a given state.

Therefore, a naive extension of diffusion policies to the multi-agent setting (i.e., training a joint policy from multi-agent expert demonstrations) faces two key challenges:
\begin{enumerate}[label=(\roman*),leftmargin=2em]
  \item the joint state-action space grows exponentially with the number of agents, making it difficult to ensure good coverage of the entire space with expert demonstrations,
  \item providing demonstrations for multi-agent systems is cumbersome, even simply due to the logistical challenge of controlling $N$ agents simultaneously.
\end{enumerate}
To address these challenges, this section introduces our \textbf{main contribution}: \emph{a framework for composing multi-agent behavior from single-agent diffusion policies.}

\para{Notation: Single-Agent vs. Multi-Agent Quantities}
Before we proceed to our main contribution, we introduce the following notation to distinguish between single-agent and multi-agent quantities. We use superscripts, i.e., $(\cdot)^{(i)}$, to denote quantities relating to the $i$-th of $N$ agents and omit such superscripts when we refer to quantities pertaining to all $N$ agents jointly.
Following this convention, we use
\begin{itemize}[leftmargin=2em]
  \item $\state^{(i)}\in\statespace^{(i)}$ to denote the state of the $i$-th agent, and $\jointstate\in\jointstatespace$ to denote the joint state of all $N$ agents, both of which are related via the mapping $\statedecomposer{i}: \jointstatespace\to\statespace^{(i)}$ so that $\state^{(i)} = \statedecomposer{i}(\jointstate)$,
  \item $\action^{(i)}\in\actionspace^{(i)}$ to denote the action of the $i$-th agent, $\jointaction\defeq(\action^{(1)}, \ldots, \action^{(N)})\in\jointactionspace\defeq\prod_{i=1}^N\actionspace^{(i)}$ to denote the joint action of all $N$ agents,
  \item $\mapolicy(\jointaction \given \jointstate)$ to denote the multi-agent policy that characterizes a distribution over joint actions given a joint state,
  \item and $\dataset^{(i)} = \{(\state_{(d)}^{(i)}, \action_{(d)}^{(i)})\}_{d=1}^{D_i}$ to denote a dataset of single-agent demonstrations for agent $i$.
\end{itemize}

\subsection{Key Idea: Multi-Agent Learning with Single-Agent Data Constraints}
\label{sec:problem-setup}
Consider the scenario shown in~\cref{fig:method-overview}.
Here, two robotic arms are separated on a wide table and must move an object from one location to another. 
Since each robot can only reach a limited area of the table, they need to collaborate to complete the task.
Nevertheless, several skills required to accomplish this task are relevant even to a single agent, like the ability to pick up objects and place them elsewhere. 
Once both robots possess these skills, more complex behaviors---such as passing an object to another agent who cannot otherwise reach the object---can emerge through careful \emph{coordination} of individual skills across agents.

With this intuition, we propose \acf{codi} which consists of three key steps (\cref{fig:method-overview}).
\begin{description}[leftmargin=1em]
  \item[Step 1: Single-Agent Pretraining.] The user provides single-agent demonstrations $\dataset^{(i)}$ of basic skills for each agent (in our running example, picking and placing objects), from which we fit a \inlinecolorbox{intent-blue!\eqnhighlightshade}{\emph{single-agent diffusion policy} $\pdiffusion^{(i)}$} (hereafter, \emph{base policy}) capturing a probabilistic \say{library} of base-skills. When agents are homogeneous, the same policy can be shared across agents.
  \item[Step 2: Guidance Design.] The user designs a \inlinecolorbox{intent-teal!\eqnhighlightshade}{\emph{multi-agent cost function} $\guide$}: $\jointstatespace\times\jointactionspace \to\reals$, that penalizes deviations from desired multi-agent behavior (in our running example, rewarding object movement towards the goal location while penalizing collisions with other agents).
  \item[Step 3: Deployment via Multi-Agent Policy Composition.] At test time, we generate coordinated multi-agent behavior by composing the single-agent diffusion policies with the multi-agent cost function into a multi-agent diffusion policy, $\madiffusionpolicy$.
\end{description}

Mathematically, \ac{codi} uses the ingredients from Step~1 (the single-agent base policies $\pdiffusion^{(i)}$) and Step~2 (the multi-agent cost function $\guide$) to solve the following KL-regularized \emph{cooperative game} in Step~3: 
\vspace{1em}
\begin{align}
  \label{eq:joint-optimization-problem}
  \madiffusionpolicy(\jointaction\given\jointstate) \in \arg\min_{\tilde\policy\in\mapolicyspace}
  \expectation_{\jointaction\sim\tilde\policy}
  \left[
    \eqnmarkbox[intent-teal]{b1b}{\guide}(\jointaction, \jointstate)
    \right]
  - \oursregularizationfactor\kldiv{\tilde\policy}{\eqnmarkbox[intent-blue]{b4}{\madatadriven}},\nonumber \\
  \text{where }\eqnmarkbox[intent-blue]{b3}{
    \madatadriven(\jointaction\given{\jointstate}) \defeq \prod_{i=1}^N \pdiffusion^{(i)}
  }
  (\action^{(i)} \given \statedecomposer{i}(\jointstate)),
  \annotate[yshift=0.1em]{below,left}{b3}{\small learned from single-agent data}
  \annotate[yshift=0.4em]{above,left}{b1b}{user-defined multi-agent cost}
\end{align}%
Intuitively, in this game, agents \emph{jointly} seek a policy $\tilde\policy\in\mapolicyspace$ from the space of joint policies that minimizes their expected cost under $\guide$ while staying close—in KL-divergence—to the product of their independently trained base policies $\madatadriven$.

Via the Donsker-Varadhan variational formula~\cite{donsker1975asymptotic}, %
this cooperative game has the unique closed-form solution:
\begin{align}
  \label{eq:codi-structure}
  \madiffusionpolicy(\jointaction \given \jointstate) \defeq
  \frac{1}{\oursnormalizingconstant(\jointstate)}
  \exp\left(\frac{-\eqnmarkbox[intent-teal]{b1a}{\guide}(\jointstate, \jointaction)}{\oursregularizationfactor}\right)
  \eqnmarkbox[intent-blue]{b2}{
    \madatadriven
  }
  (\jointaction\given\jointstate).
\end{align}
Here, $\madatadriven(\jointaction\given{\jointstate})$ takes the role of a prior, structured as the product of (independent) single-agent diffusion policies, and $\exp\left({-\guide(\jointstate, \jointaction)}{\oursregularizationfactor}^{-1}\right)$ takes the role of a coupling term, assigning higher probability to joint actions that result in lower cost, with $\oursregularizationfactor\in\reals_{\geq0}$ controlling the strength of this coupling and the scalar $\oursnormalizingconstant(\jointstate)\defeq\int_{\jointactionspace}\exp\left({-\guide(\jointstate, \jointaction)}{\oursregularizationfactor}^{-1}\right)\intd\jointaction$ being the normalization constant.

The structure of \cref{eq:codi-structure} naturally raises two further questions: Does the product prior $\madatadriven$ impose an undesirable independence assumption across agents? And what class of multi-agent behaviors can this structure actually represent? The following sections address each of these questions in turn.

\subsection{Can \ac{codi} Capture Interdependent Behavior?}
\label{sec:sklars}
A natural concern is that the product prior $\madatadriven$ might enforce an undesirable independence assumption across agents.
A useful lens for this issue comes from \emph{copula} theory~\cite{nelsen2006introduction}, which studies how to build joint distributions from marginal models while explicitly modeling dependence.
Its core result, Sklar's theorem~\cite{sklar1959fonctions}, states that any joint density $\rho(z_1,\ldots,z_d)$ with continuous marginals $\rho_j(z_j)$ admits the unique factorization
\begin{align}
  \label{eq:sklar}
  \rho(z_1, \ldots, z_d) = c\bigl(F_1(z_1), \ldots, F_d(z_d)\bigr) \cdot \prod_{j=1}^d \rho_j(z_j),
\end{align}
where $F_j(z_j) \defeq \int_{-\infty}^{z_j} \rho_j(z') \intd z'$ is the $j$-th marginal cumulative distribution function and $c: [0,1]^d \to (0, \infty)$ is the \emph{copula density}, which encodes the full dependence structure of the joint distribution independently of the marginals.\footnote{While the original Sklar's theorem has been developed for \emph{scalar} marginals, extensions to the multivariate case, so called \emph{vector copulas}, exist~\cite{fan2023vector}.}

The proposed policy structure $\madiffusionpolicy$ in~\cref{eq:codi-structure} mirrors this form, but replaces the true marginals of the target joint distribution with user-demonstrated single-agent policies.
Below, we characterize the conditions under which this substitution still allows us to recover a desired multi-agent behavior.

\subsection{What Multi-Agent Behaviors Can We Express with Single-Agent Base Policies?}
\label{sec:design-space}
Let us take a moment to understand what kind of multi-agent behavior we can capture via the proposed policy structure of \cref{eq:codi-structure}.
The key question guiding this analysis is: \emph{where does the \say{knowledge} of the multi-agent behavior really come from?}
Is it baked into the single-agent demonstrations provided by the user? Or is it encoded in the user-defined multi-agent cost function?

To formalize this analysis, let $\matargetpolicy$ denote a \emph{hypothetical multi-agent target policy} that we wish to approximate with $\madiffusionpolicy$ via our proposed approach.
Note that this target policy can be of arbitrary structure, i.e., it can be any distribution over the joint action space $\jointactionspace$; not limited to the composite form of \cref{eq:codi-structure}.
Under which conditions can our learned policy, $\madiffusionpolicy$, approximate such a target policy $\matargetpolicy$?

\para{Decomposing the Approximation Error}
To answer this question, we can decompose the \ac{kl}-divergence between the target policy $\matargetpolicy$ and our learned policy $\madiffusionpolicy$ as follows:
\begin{subequations}
  \label{eq:kl-decomposition}
  \begin{align}
    \label{eq:kl-decomposition-initial}
     & \kldiv{\matargetpolicy}{\madiffusionpolicy}                              \\
     & \quad = \expectation_{\jointaction\sim\matargetpolicy}\left[
      \eqnmarkbox[gray]{b6}{\log\matargetpolicy(\jointaction\given\jointstate)} -
      {\log\madiffusionpolicy(\jointaction\given\jointstate)}
    \right]\nonumber                                                            \\
    \label{eq:kl-decomposition-decomposed}
     & \quad = \underbrace{\expectation_{\jointaction\sim\matargetpolicy}\left[
        \log\frac{\eqnmarkbox[gray]{b6}{\matargetpolicy(\jointaction\given\jointstate)}}{
          \eqnmarkbox[intent-blue]{b6}{\madatadriven(\jointaction\given\jointstate)}}
        \right]}_{= \kldiv{\matargetpolicy}{\madatadriven} \geq 0}
    +
    \underbrace{\eqnmarkbox[intent-teal]{b6}{
        \log\oursnormalizingconstant +
        \expectation_{\jointaction\sim\matargetpolicy}\left[
          \frac{\guide(\jointstate,\jointaction)}{\oursregularizationfactor}
          \right]
      }}_{\neq 0}.
  \end{align}
\end{subequations}
Here, the first corresponds to the \ac{kl}-divergence between the \inlinecolorbox{gray!\eqnhighlightshade}{target policy $\matargetpolicy$} and the \inlinecolorbox{intent-blue!\eqnhighlightshade}{data-driven component $\madatadriven$}, and the second corresponds to the contribution of the \inlinecolorbox{intent-teal!\eqnhighlightshade}{user-defined multi-agent cost function $\guide$}.

\para{Analyzing the Approximation Error}
From this decomposition, it is clear that, if the data-driven component $\madatadriven$ (learned from single-agent demonstrations) does not capture the target behavior $\matargetpolicy$ perfectly, i.e., $\kldiv{\matargetpolicy}{\madatadriven}>0$, then the multi-agent cost function $\guide$ must compensate for this error.
In \cref{appendix:optimally-compensating-multi-agent-cost-function} we analyze this case, revealing that error compensation is achievable when the target behavior~$\matargetpolicy$ is absolutely continuous w.r.t. the data-driven component~$\madatadriven$ ($\matargetpolicy \gg \madatadriven$); i.e., $\madatadriven(\jointaction\given\jointstate)=0$ must imply $\matargetpolicy(\jointaction\given\jointstate)=0$ for all $\jointaction\in\tilde\jointactionspace\subseteq\jointactionspace$ for which $\tilde\actionspace$ has non-zero measure under $\matargetpolicy$.
Intuitively, this means that the product of single-agent policies must be an over-approximation of the target multi-agent behavior.
Given such an over-approximation, the role of the cost function then is to promote desired multi-agent behavior from the product of single-agent policies while suppressing any joint actions that amount to coordination failures (e.g.,\ collisions).
Assuming absolute continuity, the optimally compensating multi-agent cost function is given by
\begin{align}
  \label{eq:optimal-guide}
  \guide^*(\jointstate,\jointaction) & = -\oursregularizationfactor\log\oursnormalizingconstant(\jointstate)
  +\oursregularizationfactor\log\frac{\madatadriven(\jointaction\given\jointstate)}{\matargetpolicy\left(\jointaction\given\jointstate\right)}.
\end{align}
This result highlights that the role of $\guide$ is to account for the log-likelihood gap between $\matargetpolicy$ and $\madatadriven$.
If the demonstrations are perfectly informative, we would have $\log\frac{\madatadriven(\jointaction\given\jointstate)}{\matargetpolicy\left(\jointaction\given\jointstate\right)} = 0$ and it is easy to see that $\guide^*(\jointstate,\jointaction) \propto 1$.
The larger the \ac{kl}-gap between $\madatadriven$ and $\matargetpolicy$ is, the more information must be encoded in the cost function to compensate for this error.

\para{Key Insight}
In conclusion, this analysis reveals two important results when trying to approximate multi-agent behavior $\matargetpolicy$ with $\madiffusionpolicy$ in the proposed form of \cref{eq:codi-structure} with our method.
First, it reveals an important requirement for the single-agent demonstrations provided by the user:
\emph{
  The single-agent demonstrations must be sufficiently rich to cover the support of the target behavior $\matargetpolicy$.
}
And second, \cref{eq:optimal-guide} reveals that the more informative the single-agent demonstrations are for the targeted multi-agent behavior, the less design effort needs to go into $\guide$.

\section{Sampling from \ac{codi} Policies}
So far, we have established \emph{what} policy structure \ac{codi} uses, and analyzed when this structure can accurately approxiamte multi-agent behavior.
The remaining question is \emph{how} to generate samples from this policy so that agents can actually act at test time.
Sampling from the multi-agent policy in \cref{eq:codi-structure} is non-trivial for several reasons.
First, even computing the correct normalizing constant is challenging.
Moreover, since the data-driven component is represented by diffusion policies, its distribution is only known in terms of a generative model: generating samples is easy but computing the density function is not (cf. \cref{sec:background}).

In this section, we show how to exploit the structure of~\cref{eq:codi-structure} to directly construct a generative model for the multi-agent policy, thereby tackling the challenges of sampling from the multi-agent policy.
We present this key result in two steps.
First, in~\cref{sec:constructing-a-generative-model-for-the-multi-agent-policy}, we derive a reverse-time \ac{sde} following the structure of~\cref{eq:reverse-sde} whose numerical integration generates samples from the multi-agent policy~$\madiffusionpolicy(\jointaction\given\jointstate)$.
Then, in~\cref{sec:multi-agent-guidance-score-estimation}, we show how to construct estimators for the required ingredients of this generative model.

\subsection{Constructing a Generative Multi-Agent Policy}
\label{sec:constructing-a-generative-model-for-the-multi-agent-policy}
To construct a generative model for the multi-agent policy, we seek a \ac{sde} of the form of~\cref{eq:reverse-sde} whose solution matches the target distribution $\madiffusionpolicy(\jointaction\given\jointstate)$.

\para{The Multi-Agent Reverse-Time SDE}
In \cref{sec:background-reverse-diffusion}, we saw that a generative diffusion model for a distribution $\ptarget(\sample)$ can be constructed by learning a score model that satisfies $\scoremodel(\sample; t) \approx \truescore(\sample; t)$ and solving the reverse-time \ac{sde} in~\cref{eq:reverse-sde-with-score-model} backwards in time from $T$ to $0$.
By symmetry, in order to construct a generative model for $\madiffusionpolicy(\jointaction\given\jointstate)$, we need a score model that satisfies $\mascoremodel(\jointaction; t, \jointstate) \approx \nabla_{\jointaction} \log \madiffusionpolicy_{t}(\jointaction;\jointstate)$, where $\madiffusionpolicy_{t}(\jointaction;\jointstate) \defeq \int_{\jointactionspace} \madiffusionpolicy(\jointaction\given\jointstate) \perturbationkernel{\jointaction} \intd{\jointaction}$ and solve
\begin{align}
  \label{eq:multi-agent-reverse-sde}
  \intd \jointaction(t) & = -2t\overbrace{\mascoremodel(\jointaction(t); t, \jointstate)}^{\approx \nabla_{\jointaction(t)} \log \madiffusionpolicy_{t}(\jointaction(t);\jointstate)} \intd t + \sqrt{2t} \intd \wienerback_t\nonumber \\
  \jointaction(T)       & \sim \underbrace{\diffusionprior}_{\approx \madiffusionpolicy_{T}(\jointaction(T);\jointstate)}.
\end{align}
The only unknown in this equation is the \emph{multi-agent score model} $\mascoremodel(\jointaction; t, \jointstate)$.
Unlike in conventional diffusion models, however, we cannot trivially learn this score model via \ac{dsm} from samples as in \cref{eq:score-matching-loss-simplified} because we precisely do \emph{not} have access to any samples from the multi-agent target policy. %
Next, we discuss how to overcome this challenge by directly constructing the multi-agent score $\mascoremodel(\jointaction; t, \jointstate)$ from known ingredients.

\para{Recognizing the Single-Agent Score Contribution to the Multi-Agent Score}
To find an estimator of the multi-agent score model, we first decompose it in terms of the single-agent score models.
To this end, as we detail in Appendix~\ref{appendix:decomposition-of-madiffusionpolicy}, we can decompose the multi-agent score as follows:
\begin{align}
  \label{eq:multi-agent-score}
   & \nabla_{\jointaction(t)} \log \madiffusionpolicy_{t}(\jointaction(t);\jointstate)
  =                                                                                            \\
   & ~\qquad
  \eqnmarkbox[intent-blue]{b7}{
  \nabla_{\jointaction(t)} \log \madatadriven[t](\jointaction(t)\given\jointstate)
  }\nonumber                                                                                   \\
   & \quad +
  \underbrace{\eqnmarkbox[intent-teal]{b8}{
      \nabla_{\jointaction(t)} \log \expectation_{\jointaction\sim \madatadriven(\jointaction\given \jointaction(t), \jointstate, t)}\left[
        \exp\left(\frac{-\guide(\jointstate,\jointaction)}{\oursregularizationfactor}\right)
        \right]
    }}_{\guidancescore(\jointaction(t);\jointstate,t)\defeq}.
  \nonumber
\end{align}
Analyzing the \inlinecolorbox{intent-blue!\eqnhighlightshade}{first term} reveals that it is the score of the data-driven component $\madatadriven$ whose score model can be trivially constructed from the scores learned by the single-agent diffusion policies:
\begin{subequations}
  \begin{align}
    \label{eq:single-agent-score}
     & \nabla_\jointaction \log \madatadriven[t](\jointaction(t)\given\jointstate)\nonumber                                               \\
     & \quad =
    \nabla_\jointaction \log \prod_{i=1}^N \pdiffusion[t]^{(i)}(\action^{(i)}\given\statedecomposer{i}(\jointstate))                    \\
     & \quad =
    \begin{bmatrix}
      \nabla_{\action^{(1)}} \log \pdiffusion[t]^{(1)}(\action^{(1)}\given\statedecomposer{1}(\jointstate)) \\
      \vdots                                                                                                           \\
      \nabla_{\action^{(N)}} \log \pdiffusion[t]^{(N)}(\action^{(N)}\given\statedecomposer{N}(\jointstate))
    \end{bmatrix} \\
     & \quad=
    \begin{bmatrix}
      \scoremodel[(1)](\action^{(1)}; t, \statedecomposer{1}(\jointstate)) \\
      \vdots                                                                           \\
      \scoremodel[(N)](\action^{(N)}; t, \statedecomposer{N}(\jointstate))
    \end{bmatrix}                          \\
     & \quad=: \productscoremodel(\jointaction; t, \jointstate), \nonumber
  \end{align}
\end{subequations}
where $\scoremodel[(i)]$ denotes the score model underpinning the $i$-th agent's single-agent diffusion policy $\pdiffusion^{(i)}$.
This reveals that we do not need to approximate the multi-agent score from scratch: we can re-use the single-agent score models already available from the underpinning base policies.
Consequently, the only missing ingredient is the \inlinecolorbox{intent-teal!\eqnhighlightshade}{second term} in \cref{eq:multi-agent-score} which encodes the contribution of the user-defined multi-agent cost function $\guide$.
As we show in \cref{appendix:relationship-to-conventional-classifier-guidance}, this term can be related to the framework of classifier guidance~\cite{dhariwal2021diffusion}. We therefore refer to \inlinecolorbox{intent-teal!\eqnhighlightshade}{$\guidancescore(\jointaction(t);\jointstate,t)$} as the \inlinecolorbox{intent-teal!\eqnhighlightshade}{guidance score}.

\para{Key Insight}
We can sample from our multi-agent policy~$\madiffusionpolicy$ in \cref{eq:codi-structure} by numerically integrating the reverse diffusion~\ac{sde} in~\cref{eq:multi-agent-reverse-sde}.
The score model involved in this \ac{sde} can be constructed from the \inlinecolorbox{intent-blue!\eqnhighlightshade}{single-agent scores} (which we already have from pre-training the single-agent diffusion policies) plus an extra \inlinecolorbox{intent-teal!\eqnhighlightshade}{guidance score term}.
Constructing an estimator for this guidance score term will be the focus of the next section.

\subsection{Multi-Agent Guidance Score Estimation}
\label{sec:multi-agent-guidance-score-estimation}

Having established that the only missing ingredient required for sampling from the multi-agent policy is the guidance score, $\guidancescore(\jointaction(t);\jointstate,t)$, in~\cref{eq:multi-agent-score}, we construct an estimator for this term.

\para{Design Constraints}
When constructing this estimator, we specifically consider two design constraints: (\romannumeral1)~the user-provided cost function $\guide$ may not be differentiable; and (\romannumeral2)~we do not have access to samples from the multi-agent target policy $\madatadriven(\jointaction\given\jointstate)$.
Hence, this estimator must be constructed from point-wise evaluations of the black-box cost $\guide$, without access to its derivatives.

\subsubsection{Sampling-Based Score Estimation}
\label{sec:online-guidance-score-estimation}
Our guidance score estimator approximates $\guidancescore(\jointaction(t);\jointstate,t)$ in a sampling-based (i.e., gradient-free) manner.
To achieve this, as shown in Appendix~\ref{appendix:simplification-of-guidancescore}, we rewrite $\guidancescore(\jointaction(t);\jointstate,t)$ as follows:
\begin{align}
  \label{eq:guidance-score-estimator-gradient-free}
  &\guidancescore(\jointaction(t);\jointstate,t) =\\
  &\quad\expectation_{\jointaction\sim \madatadriven(\jointaction\given \jointaction(t), \jointstate, t)}\left[
    \weight(\jointaction(t), \jointaction, \jointstate, t) \nabla_{\jointaction(t)} \log \perturbationkernel{\jointaction}
    \right]\nonumber
\end{align}
where
\begin{align}
  \weight(\jointaction(t), \jointaction, \jointstate, t) \defeq
  \frac{
    \exp\left(\frac{-\guide(\jointaction;\jointstate)}{\oursregularizationfactor}\right)
  }{
    \expectation_{\jointaction\sim \madatadriven(\jointaction\given \jointaction(t), \jointstate, t)}\left[
      \exp\left(\frac{-\guide(\jointaction;\jointstate)}{\oursregularizationfactor}\right)
      \right]
  }-1
\end{align}
This formulation side-steps the need to compute gradients of $\guide$.\footnote{
  When $\guide$ is differentiable in $\jointaction$, we can alternatively compute a lower-variance estimate of the expectation in~\cref{eq:guidance-score-estimator-gradient-free} using the reparameterization trick. Here, however, we are interested in the general case where the user-provided multi-agent cost need not be differentiable.
}
Next, we approximate the expectations in \cref{eq:guidance-score-estimator-gradient-free} in a sampling-based manner using Monte Carlo integration.
This requires tractable generation of samples from the posterior $p(\jointaction\given\jointaction(t), \jointstate, t)$ (sampling \emph{noise-free} data $\jointaction$ given \emph{noisy} data $\jointaction(t)$).
To this end, we construct a Gaussian posterior approximation using Tweedie's formula~\cite{robbins1992empirical,efron2011tweedie}
as shown in Appendix~\ref{appendix:gaussian-approximation-of-posterior}, admitting
\begin{align}\label{eq:gaussian-approximation-of-posterior}
  p(\jointaction\given \jointaction(t), \jointstate, t) &\approx
  \normal(
  \jointaction;
  \muonlineguidance,
  \sigmaonlineguidance
  ),\\
  \text{where }\muonlineguidance &\defeq \jointaction(t) + t^2 \productscoremodel(\jointaction(t); t, \jointstate), \nonumber\\
  \sigmaonlineguidance &\defeq t^2 \identity + t^4 \nabla_{\jointaction(t)} \productscoremodel(\jointaction(t); t, \jointstate) \nonumber
\end{align}
Observe that the parameters $\muonlineguidance$ and $\sigmaonlineguidance$ of this distribution are constructed directly from the single-agent score models characterizing $\productscoremodel(\jointaction(t); t, \jointstate)$ (cf. \cref{eq:single-agent-score}), requiring no additional training.
In practice, we find that the higher-order term $t^4 \nabla_{\jointaction(t)}^2 \productscoremodel(\jointaction(t); t, \jointstate)$ is negligible, side-stepping the need to compute the score-model's Jacobian at runtime.

\para{Key Insight}
In summary, exploiting~\crefrange{eq:guidance-score-estimator-gradient-free}{eq:gaussian-approximation-of-posterior} and applying Monte Carlo integration, we can approximate the guidance score of \cref{eq:guidance-score} with $\montecarlosamplesize$ samples as
\begin{align}
  \label{eq:monte-carlo-guidance-score}
  &\guidancescore(\jointaction(t);\jointstate,t) \approx
  \onlineguidancescoreestimate(\jointaction(t);\jointstate,t) \defeq\\
  &\quad\qquad\frac{1}{\montecarlosamplesize}\sum_{m=1}^{\montecarlosamplesize}
  \weight_\text{codi}^\guide(\jointaction^{(m)}, \jointstate)
  \nabla_{\jointaction(t)} \log \perturbationkernelsym_{t}(\jointaction(t)\given\jointaction^{(m)})\nonumber
\end{align}
where $\jointaction^{(m)}$ are sampled from \cref{eq:gaussian-approximation-of-posterior}
and
\begin{align}
\weight_\text{codi}^\guide(\jointaction, \jointstate) &\defeq \exp\left(\frac{-\guide(\jointstate,\jointaction)}{\oursregularizationfactor}\right)/\mu_\weight - 1,\\
\text{with }\mu_\weight &\defeq \frac{1}{\montecarlosamplesize}\sum_{m=1}^{\montecarlosamplesize}\exp\left(\frac{-\guide(\jointstate,\jointaction^{(m)})}{\oursregularizationfactor}\right).
\end{align}
This estimator can be evaluated for varying choices of $\guide$ without retraining. %

\section{Hardware Experiments}
\label{sec:hardware-results}

We begin our evaluation by demonstrating the qualitative behavior of \ac{codi}.
To this end, we instantiate the out-of-reach manipulation task from \cref{fig:method-overview} in hardware on two 7-DoF Franka Research 3 arms.
These hardware experiments aim to demonstrate the practical applicability of our method and the collaborative strategies it discovers.

\begin{figure*}[t]
    \centering
    \captionsetup{skip=0pt}
    \includegraphics[width=0.198\textwidth]{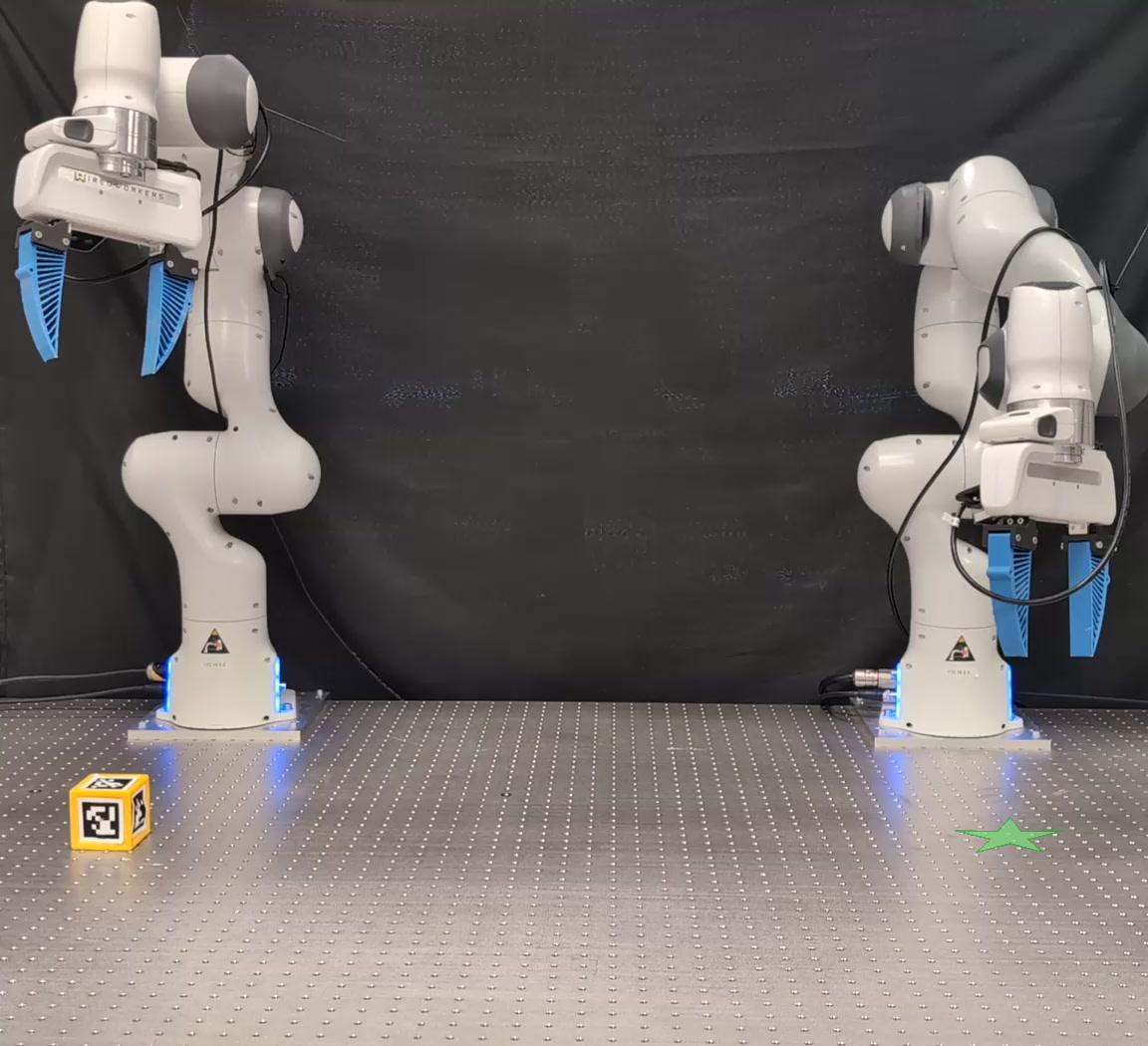}\hfill
    \includegraphics[width=0.198\textwidth]{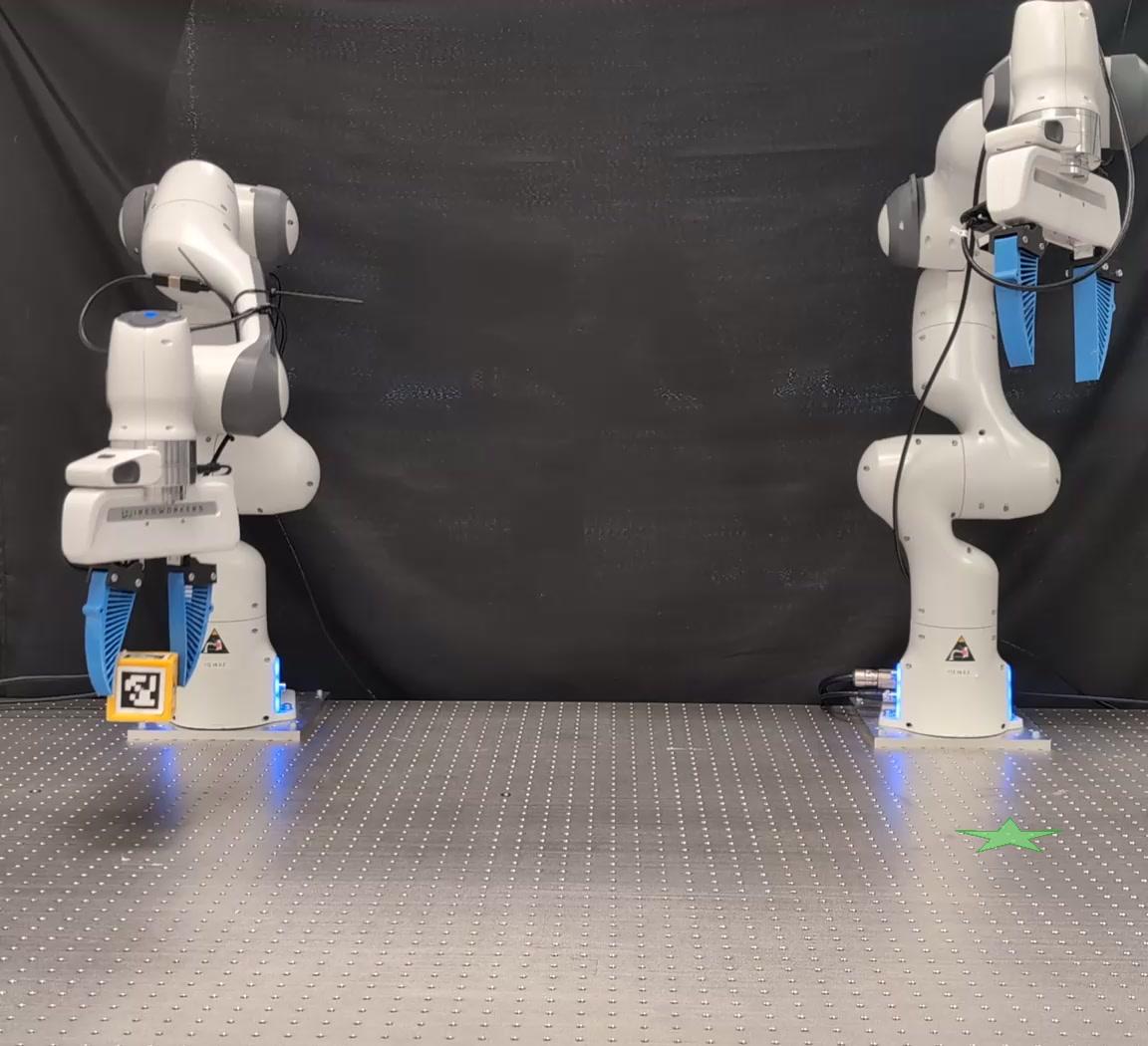}\hfill
    \includegraphics[width=0.198\textwidth]{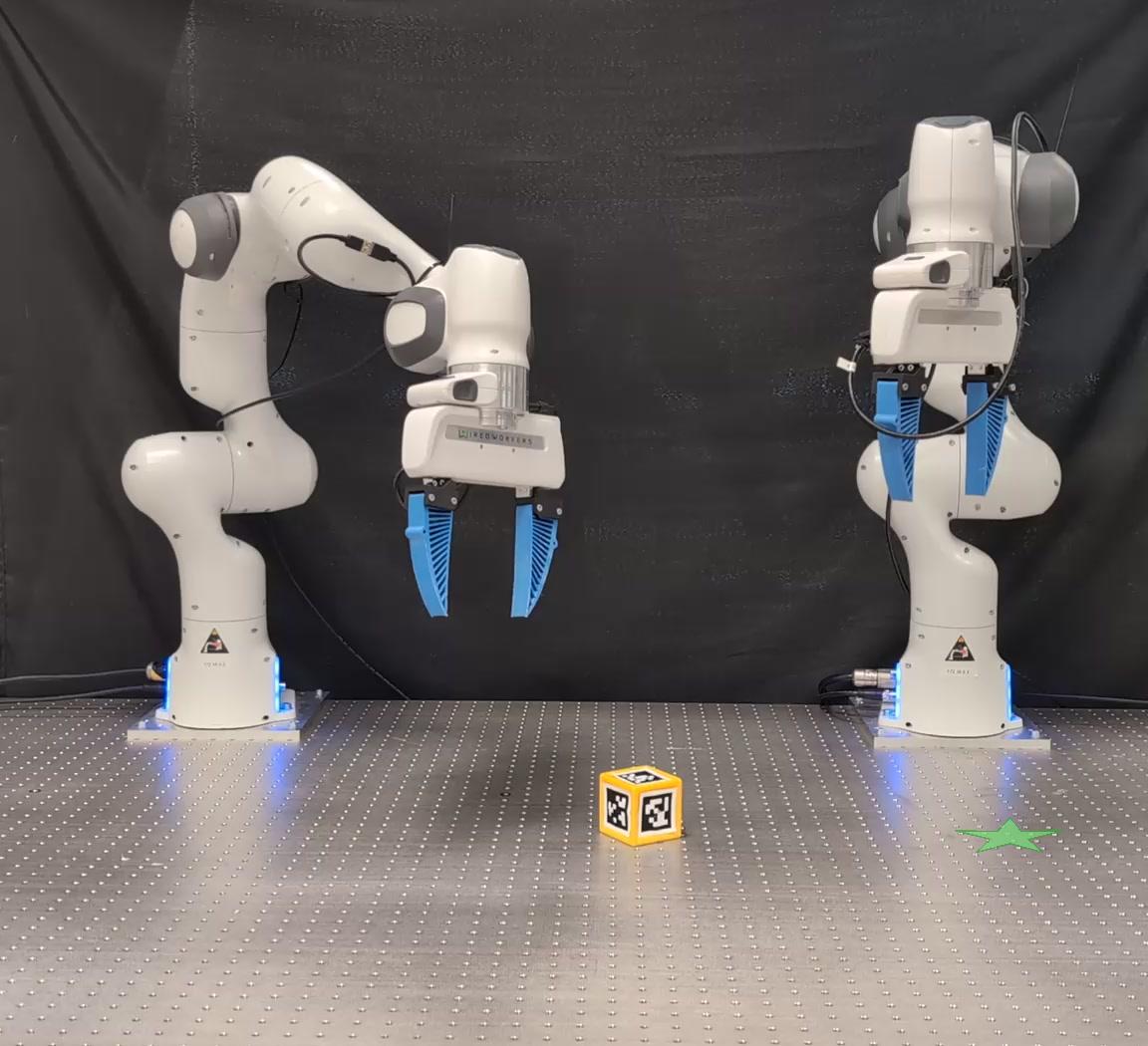}\hfill
    \includegraphics[width=0.198\textwidth]{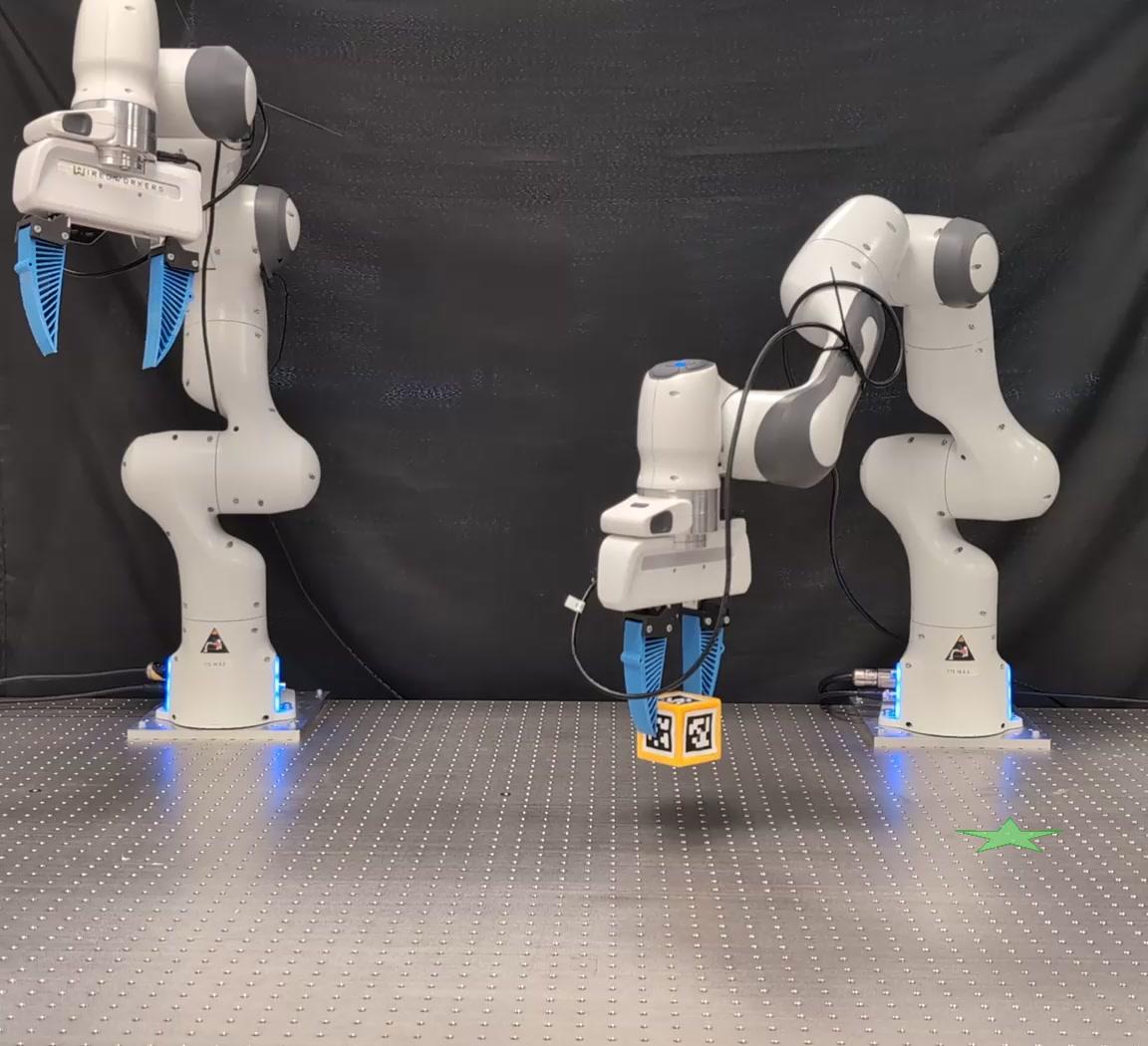}\hfill
    \includegraphics[width=0.198\textwidth]{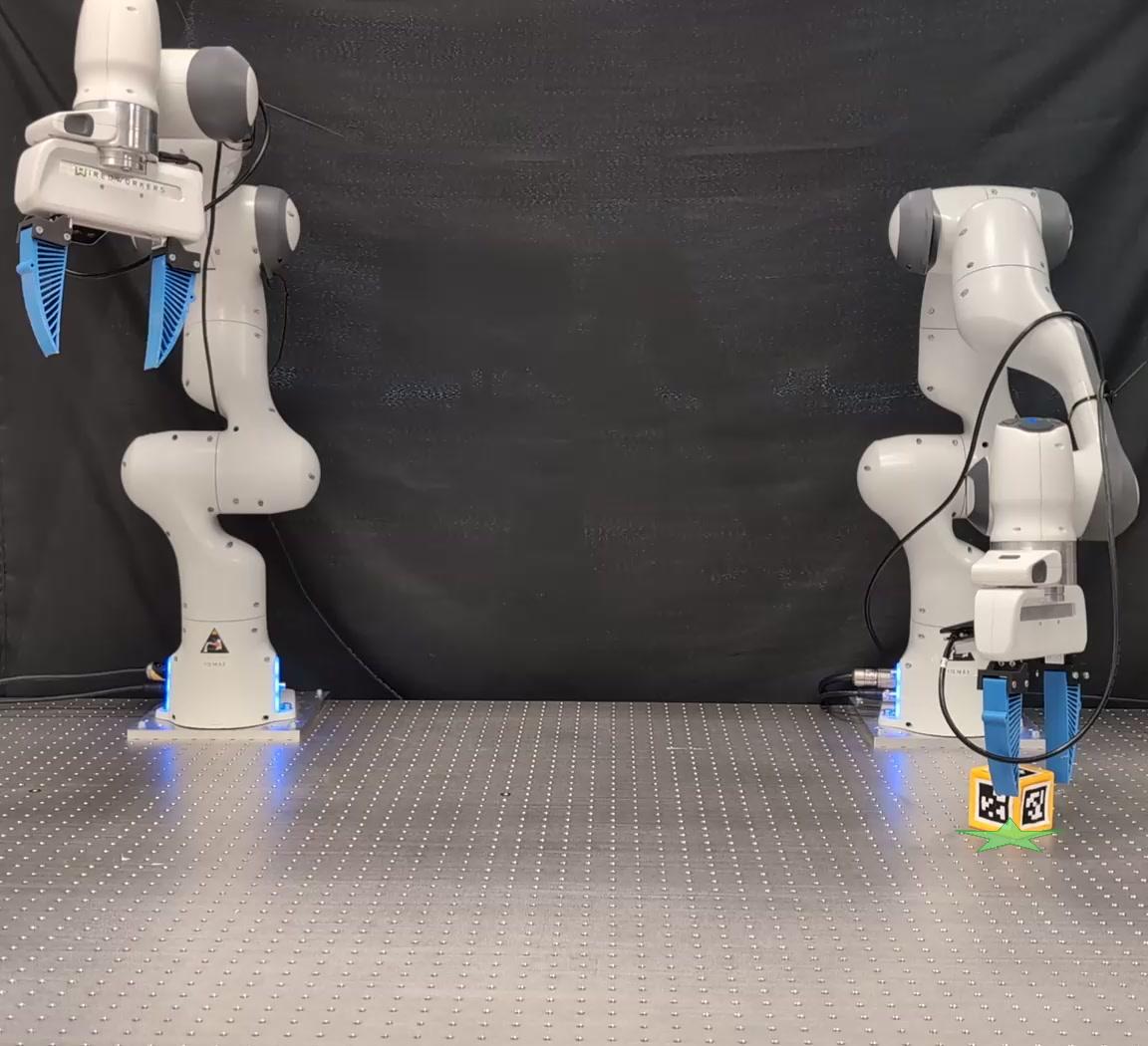}\hfill
    \caption{Qualitative example of multi-agent behavior generated by \ac{codi}.
    Our approach guides the left robot to pick up the cube and place it at an intermediate location reachable by the right robot, which places it at the goal marked by the green star.
    }
    \label{fig:hardware-example}
\end{figure*}

\para{Task}
The robots are tasked with moving the object (a cube with an edge length of \SI{5}{cm}) to the goal location (marked by a green star in \cref{fig:hardware-example}) within a tolerance of \SI{15}{cm}.
Throughout the experiment, we repeatedly reset the cube to an arbitrary position on the left side of the workspace.
Note that the dimensions of the table (width \SI{1.8}{m}, depth \SI{1.2}{m}) are such that no single robot can reach across the entire table.
Therefore, both robots must collaborate to complete the task, e.g., place the object at an intermediate location reachable by the other robot, which then completes the task by placing the object at the goal.

\begin{figure}[t]
    \centering
    \captionsetup{skip=0pt}
    \subfloat[Full training data.\label{fig:initial_configurations_full}]{%
        \includegraphics[scale=1.0]{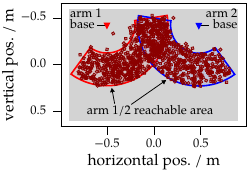}}%
    \hfill
    \subfloat[25\% training data.\label{fig:initial_configurations_quarter}]{%
        \includegraphics[scale=1.0]{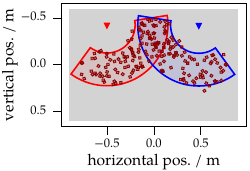}}%
    \caption{Distribution of initial cube positions used for training.
    }
    \label{fig:initial_configurations}
\end{figure}

\para{Demonstration Data}
We train our method entirely in simulation by replicating our hardware setup in the high-fidelity Isaac Gym simulator~\cite{makoviychuk2021isaac}.
We collect pick-and-place demonstrations using a scripted controller that drives the end-effector via inverse kinematics to randomly sampled goal positions.
The dataset is split equally between \emph{pick} demonstrations, where the robot interacts with reachable cubes, and \emph{yield} demonstrations, where the robot returns to its home pose to leave room for the other arm at test time.
Finally, to learn recovery behaviors, we reset the cube to a random position on the table at each time step with a 1\% probability, prompting the expert controller to re-grasp the cube if it is within reach.
A video example of these demonstrations is available in the supplementary material.
The set of initial cube positions used for training is visualized in \cref{fig:initial_configurations_full}; goal positions are sampled from the same distribution.
The resulting dataset comprises 1k demonstrations, split into 500k receding-horizon segments of 16 action predictions at a rate of \SI{10}{Hz}.

\para{Cost Function Design}
The demonstration data only encodes knowledge of the base \emph{skills} (here, the ability to pick up and place the cube); not of the exact task to be performed (here, the position where the cube should be placed).
Instead, the task-specific information is encoded in the cost function.
This setup targets a scenario where the user may want to use the same base policy for multiple different tasks, side-stepping the need to provide task-specific demonstrations. This philosophy is consistent with the single-agent approach of~\citet{janner2022planning}.

The cost function used for guidance takes the form of a weighted sum of cost components,
\begin{align}
    \label{eq:franka-cost-function}
    \guide(\jointstate,\jointaction) = \weightedcostcomponent[goal](\jointstate,\jointaction) & + \weightedcostcomponent[collision](\jointstate,\jointaction) \\&+ \weightedcostcomponent[engage](\jointstate,\jointaction)\nonumber.
\end{align}
Here, $\costcomponent[goal](\jointstate,\jointaction)$ measures the distance of the object from the goal resulting from applying action-sequence $\jointaction$ from state $\jointstate$; $\costcomponent[collision](\jointstate,\jointaction)$ is a binary indicator, equal to one if end-effectors are closer than a given safety distance as a result of taking the joint action sequence $\jointaction$ from state $\jointstate$, and zero otherwise; and
$\costcomponent[engage](\jointstate,\jointaction)$ penalizes the distance of the closer robot to the object.
\Cref{appendix:cost-design-details} provides additional implementation details on this cost function.

Note that, in order to evaluate $\costcomponent[goal]$, we need to make a prediction about the \emph{combined effect} of all agents' actions on the future position of the object.
Since such a prediction is not part of the diffusion policies' output, we obtain the prediction by \emph{simulating} the outcome of the joint action sequence $\jointaction$ from state $\jointstate$ using Isaac Gym.
Note that this simulation can be done for many samples in parallel during guidance score estimation of~\cref{eq:monte-carlo-guidance-score}.

\para{Policy Representation}
We represent the score model underpinning single- and multi-agent diffusion policies via a denoising UNet architecture as in~\citet{chi2023diffusion-rss}.
The single-agent diffusion policy observes the ego-robot's end-effector pose, velocity, and gripper width together with the cube pose, and predicts desired end-effector translational and rotational velocities along with gripper width.
Policy outputs are converted to joint velocities via inverse kinematics.

\para{Additional Implementation Details}
We attach ArUco markers to the cube, which are tracked via a wrist-mounted camera on each arm.
Combined with the forward kinematics of both arms, this yields a complete state observation.

\para{Results: On Collaboration Strategies Discovered by \ac{codi}}
The supplementary material includes a 12-minute hardware video in which a human repeatedly places the object on the left side of the workspace while the robots are tasked with moving it to the right.
\Cref{fig:hardware-example} shows representative video frames of this qualitative evaluation.
\Ac{codi} composes the independent single-agent policies into a cohesive multi-agent policy: the left robot picks up the object and places it at an intermediate location within reach of the right robot, which then completes the task.
The attached video also includes qualitative examples in which we replace the cube with other objects, demonstrating a degree of robustness of the learned behavior.

\para{Key Takeaways}
In sum, these results support the claims that \textbf{(C1)}~\ac{codi} discovers collaboration strategies despite pre-training only on single-agent demonstrations, and that \textbf{(C2)}~\ac{codi} is sufficiently robust for hardware deployment.

\section{Simulation Experiments}
\label{sec:simulation_results}
\cref{sec:hardware-results} demonstrated the qualitative capabilities of our approach.
We now turn to simulation for a more systematic \emph{quantitative} evaluation of our approach, including comparison to both multi-agent and single-agent baselines and analysis of key design parameters.

\subsection{Experimental Setup}
\para{Task}
Our simulation task matches the hardware setup of \cref{sec:hardware-results}: two simulated 7-DoF Franka arms must move a cube to a goal position at the right end of the table, with initial cube positions sampled uniformly across the table.
All experiments are conducted in the Isaac Gym simulator~\cite{makoviychuk2021isaac}.

\para{Data Generation}
For all methods that rely on single-agent demonstrations (including our own), we use the same scripted generation strategy as in~\cref{sec:hardware-results}.
Some of the comparisons discussed below, however, involve baselines that require multi-agent demonstration data.
To generate this data, we use a scripted expert controller that follows the same procedure as in~\cref{sec:hardware-results} but randomly prioritizes one robot when multiple robots can reach the cube.
Unless otherwise stated, for a fair comparison, we train all base policies on the same amount of data (1k demonstrations).

\para{Evaluation and Metrics}
For each method, we perform closed-loop simulations for 50 episodes with initial configurations sampled from the same distribution as visualized in~\cref{fig:initial_configurations}.
We measure \emph{manipulation accuracy} as the minimum distance of the cube to the goal and \emph{task efficiency} as the task completion time.

\subsection{How does \ac{codi}'s performance compare to methods that utilize multi-agent demonstrations?}
\label{sec:sim-vs-joint}
We first study how \ac{codi}, which uses only single-agent data, compares to methods which use multi-agent demonstration data. Since multi-agent demonstration data is qualitatively richer but arguably more expensive to collect, we also study how the size of the single-agent demonstration data influences our ability to generate multi-agent behaviors. 

\para{Multi-Agent Baselines}
We instantiate several multi-agent baselines by applying established diffusion guidance schemes to a joint base policy, effectively treating both arms as a single \say{macro-agent}.

Classifier guidance (\textit{CG-Joint}):
This method is in spirit closest to \ac{codi}: like our approach, it achieves test-time behavior synthesis via guidance.
Beyond relying on a \emph{joint} base policy, however, it differs from \ac{codi} in how the guidance signal is computed.
Following  \citet{janner2022planning}, \textit{CG-Joint} first trains a noise-conditioned cost model to predict the expected cost for a given noisy state-action pair and then uses this cost model to guide the multi-agent base policy.
The remaining three baselines are instances of fine-tuning rather than guidance.
\acl{dpmd} (\textit{DPMD-Joint}): This method applies the mirror descent algorithm of \citet{ma2025efficient} to iteratively fine-tune the diffusion policy while penalizing deviations from the previous policy iterate.
\acl{sdac} (\textit{SDAC-Joint}): This method fine-tunes the base policy by minimizing the cost while regularizing for policy entropy~\cite{ma2025efficient}.
\acl{expo} (\textit{EXPO-Joint}): Following~\cite{dong2025expo}, this method iteratively searches for local improvements around the current diffusion policy to distill them back into the diffusion policy.
\Cref{appendix:baselines} provides additional details on all baselines.

\begin{figure}
    \centering
    \captionsetup{skip=0pt}
    \includegraphics[scale=1.0]{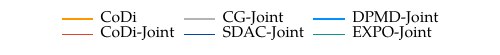}\\[-1em]
    \subfloat[Task efficiency.\label{fig:simulation-results-task-completion-times}]{%
        \includegraphics[scale=1.0]{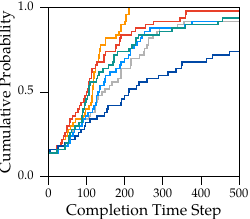}}%
    \subfloat[Manipulation accuracy.\label{fig:simulation-results-goal-distance}]{%
        \includegraphics[scale=1.0]{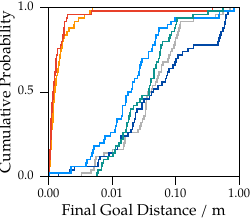}}\\[3pt]
    \caption{Comparison with baselines that utilize multi-agent demonstrations.}
    \label{fig:simulation-results-nominal}
\end{figure}

\para{Results: On the Utility of Single-Agent Pre-training}
\Cref{fig:simulation-results-nominal} summarizes the results of \ac{codi} compared to the four multi-agent baselines described above.
Despite the less informative demonstration data, we find that \ac{codi} outperforms all baselines in terms of both task efficiency and manipulation accuracy: \ac{codi} completes the task more quickly and places the cube closer to the goal than all baselines.
As an ablation, we also apply our guidance scheme directly to the \emph{multi-agent} base policy (reported as \textit{\ac{codi}-Joint} in \cref{fig:simulation-results-nominal}). We find that the performance advantage can indeed be attributed to the single-agent pre-training, as opposed to the guidance scheme itself: our guidance scheme improves the performance of the multi-agent base policy, but falls short of the performance achieved by \ac{codi} with single-agent pre-training.
This suggests that, due to the reduced size of the single-agent state-action space, single-agent pre-training gives rise to better manipulation skills than multi-agent training on the same data budget.

\begin{figure}
    \centering
    \captionsetup{skip=0pt}
    \includegraphics[scale=1.0]{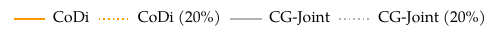}\\[-1em]
    \subfloat[Task efficiency.\label{fig:simulation-data-density-completion-times}]{%
        \includegraphics[scale=1.0]{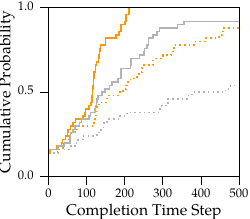}}%
    \subfloat[Manipulation accuracy.\label{fig:simulation-data-density-distance}]{%
        \includegraphics[scale=1.0]{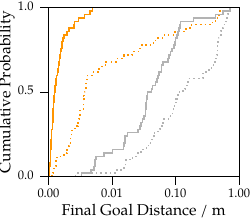}}\\[3pt]
    \caption{Performance of \ac{codi} and classifier guidance under varying demonstration data density.}
    \label{fig:simulation-data-density}
\end{figure}

\para{Results: On the Sensitivity to Demonstration Data Density}
Next, we want to understand our approach's sensitivity to the amount of demonstration data provided compared to a baseline that uses multi-agent demonstrations.
To study this, we train variants of both methods on only 20\% of the original data, i.e., 200 instead of 1k demonstrations (cf. \cref{fig:initial_configurations_quarter}), and repeat the above closed-loop evaluation.
For ease of exposition, we only compare against CG-Joint here, which was one of the strongest baselines in the above experiments.
\Cref{fig:simulation-data-density} summarizes the results of this comparison.
Despite training on only 20\% of the single-agent data, we find that \ac{codi} still achieves performance close to the full-data version of CG-Joint: it achieves a similar task success rate (88\% vs. CG-Joint's 92\% in~\cref{fig:simulation-data-density-completion-times}) while still achieving much better manipulation accuracy (cf.~\cref{fig:simulation-data-density-distance}).
By contrast, joint classifier guidance trained on only 20\% of the multi-agent demonstrations degrades to only a 54\% task success rate due to insufficient coverage of the multi-agent state-action space.

\para{Key Takeaways}
In sum, these results support the claim that \textbf{(C3)}~given the same data budget, \ac{codi} is more data-efficient than methods that rely on multi-agent demonstrations and achieves better manipulation accuracy and task efficiency at the same data budget, despite its demonstrations being cheaper to provide.


\subsection{How does joint guidance compare to independent guidance for single-agent policies?}
\label{sec:sim-vs-single}
A key aspect of our approach is that it generates coordinated \emph{joint} actions for all robots from single-agent base policies.
We seek to understand the importance of joint guidance in contrast to guiding single-agent base policies independently.

\para{Ablation: Independent Guidance}
We instantiate \textit{Codi-Indep.} as an ablation of our approach by dropping the cost components in \cref{eq:franka-cost-function} that depend on other agents.
Specifically, we drop the collision-avoidance cost and unconditionally enable the engagement cost for both agents (instead of dynamically enabling it for the closer of the two robots).

\begin{figure}
    \centering
    \captionsetup{skip=0pt}
    \includegraphics[scale=1.0]{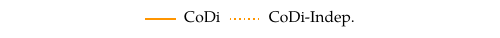}\\[-1em]
    \subfloat[Task efficiency.\label{fig:single-agent-efficiency}]{%
        \includegraphics[scale=1.0]{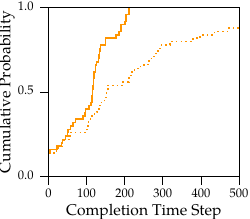}}%
    \subfloat[Manipulation accuracy.\label{fig:single-agent-accuracy}]{%
        \includegraphics[scale=1.0]{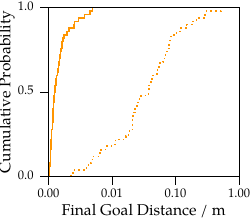}}\\[3pt]
    \subfloat[Safety.\label{fig:single-agent-safety}]{%
        \includegraphics[scale=1.0]{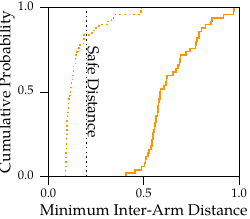}}%
    \subfloat[Failure Example.\label{fig:single-agent-failure}]{%
        \begin{minipage}[b][4.2cm][c]{0.475\linewidth}
            \centering
            \includegraphics[width=\linewidth]{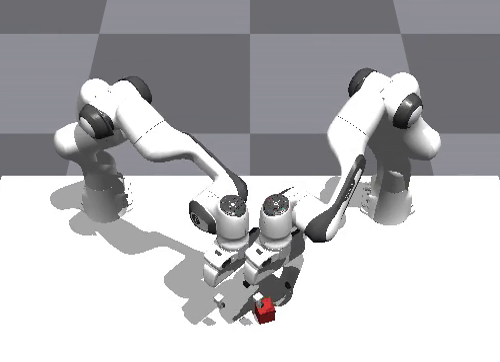}
        \end{minipage}%
    }
    \caption{
    Comparison with independent single-agent guidance.
    }
    \label{fig:single-agent-comparisons}
\end{figure}

\para{Results: On the Importance of Joint Guidance}
\Cref{fig:single-agent-comparisons} summarizes the results of evaluating this ablation in closed-loop simulation for the same 50 initial configurations as in~\cref{sec:sim-vs-joint}.
\Cref{fig:single-agent-efficiency,fig:single-agent-accuracy} show that, without joint coordination, single-agent guidance is still able to solve the task in many cases given enough time.
However, \cref{fig:single-agent-safety,fig:single-agent-failure} reveal the main failure mode of independent single-agent guidance: when agents do not coordinate their actions, both may try to pick up the cube at the same time, which leads to collisions, delays, or even complete task failures.

\para{Key Takeaways}
In sum, these results support the claim that \textbf{(C4)}~joint guidance is critical for composing independently pre-trained single-agent policies into safe and efficient coordinated multi-agent behavior.

\subsection{How should we compute the guidance signal for \ac{codi}?}
\label{sec:guidance-signal}
In \cref{sec:multi-agent-guidance-score-estimation}, we proposed a sampling-based estimator to compute the guidance score from online cost evaluations.
As noted in~\cref{sec:constructing-a-generative-model-for-the-multi-agent-policy,appendix:relationship-to-conventional-classifier-guidance}, however, this is not the only option to compute the guidance signal: the connection to classifier guidance suggests an alternative in which a multi-agent cost model is trained and its gradient used as the guidance signal.
Here we seek to understand whether the guidance signal computation matters, or whether applying classifier guidance to \ac{codi}'s product-of-marginals base policy yields comparable performance.

\para{Ablation: Classifier Guidance Applied to the Product-of-Marginals Base Policy}
To assess the importance of our sampling-based estimator, we construct an ablation that retains \ac{codi}'s product-of-marginals structure but replaces the guidance score estimate in  \cref{eq:multi-agent-score} with the gradient of a trained multi-agent cost model, following the same procedure as \textit{CG-Joint} in~\cref{sec:sim-vs-joint}.
We refer to this ablation as \textit{CG-Product}.

\begin{figure}
    \centering
    \captionsetup{skip=0pt}
    \includegraphics[scale=1.0]{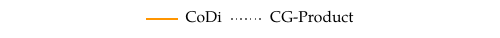}\\[-1em]
    \subfloat[Task efficiency.\label{fig:guidance-efficiency}]{%
        \includegraphics[scale=1.0]{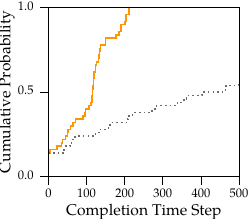}}%
    \subfloat[Manipulation accuracy.\label{fig:guidance-accuracy}]{%
        \includegraphics[scale=1.0]{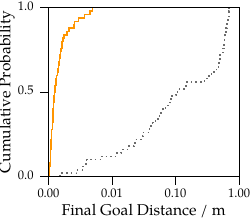}}
    \caption{
    Comparison with vanilla classifier guidance as alternative guidance instantiation for \ac{codi}.
    }
    \label{fig:guidance-comparisons}
\end{figure}

\para{Results: On the Choice of Guidance Estimator for \ac{codi}}
\Cref{fig:guidance-comparisons} shows that \textit{CG-Product} achieves substantially lower task success rates and worse manipulation accuracy than \ac{codi}.
We attribute this gap to the degree of tilt between the base policy and the target distribution.
In the joint base policy setting of \textit{CG-Joint} (\cref{sec:sim-vs-joint}), this tilt is small since joint demonstrations already encode a degree of coordination, allowing vanilla classifier guidance to perform well.
The product-of-marginals base policy, by contrast, exhibits a larger tilt relative to the coordinated target, necessitating a dedicated guidance score estimator.

\para{Key Takeaways}
This gap demonstrates that when guiding single-agent base policies, the choice of guidance estimator is critical: our sampling-based approach is necessary for effective sampling from~\cref{eq:codi-structure}~(\textbf{C5}).

\subsection{How does the support of the base policy influence guidance?}
Recall that in \cref{sec:design-space}, we discussed how guided cost sampling cannot generate behavior outside the support of the base policy.
We now empirically study this interaction between the base policy's support and the guidance cost.

\para{Methods}
To study the effect of the base policy support, we consider a simplified cost consisting solely of the goal term, $\costcomponent[goal]$ of \cref{eq:franka-cost-function}, isolating the role of the base policy from confounds introduced by a multi-term cost function.
Using this simplified cost, we evaluate two variants of \ac{codi} that differ only in their base policy choice: the first, \emph{CoDi-Simple}, uses the product of single-agent policies as in our nominal setup; the second, \emph{CoDi-Joint-Simple}, uses a joint base policy as in the multi-agent baselines of \cref{sec:sim-vs-joint}.
By construction, these base policies differ in their support structure.
The joint base policy is trained on demonstrations in which one arm always engages the cube; because it observes both end-effectors and the cube jointly, it can learn this coordination directly from data, assigning near-zero likelihood to scenarios in which neither arm is active.
In contrast, the single-agent demonstrations used to train \ac{codi}'s base policies include scenarios in which the ego robot deliberately yields to allow the other arm---unseen at training time---to handle the cube (cf.\ \cref{sec:hardware-results}).

\begin{figure}
    \centering
    \captionsetup{skip=0pt}
    \includegraphics[scale=1.0]{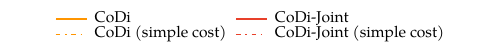}\\[-1em]
    \subfloat[Task efficiency.\label{fig:simple-cost-efficiency}]{%
        \includegraphics[scale=1.0]{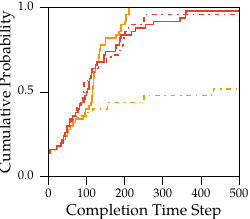}}%
    \subfloat[Manipulation accuracy.\label{fig:simple-cost-accuracy}]{%
        \includegraphics[scale=1.0]{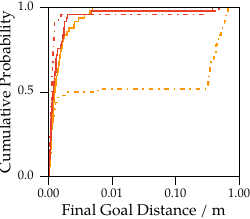}}\\[3pt]
    \caption{
        Comparison against variants that use a simplified cost model.
    }
    \label{fig:simple-cost-comparisons}
\end{figure}

\para{Results: On the Interaction Between Base Policy Support and Guidance Cost}
As shown in \cref{fig:simple-cost-comparisons}, \ac{codi} with single-agent base policies degrades substantially under the simplified cost, underscoring the critical role of the cost in composing single-agent policies into coordinated multi-agent behavior.
In contrast, replacing the base policy with one trained on joint demonstrations yields consistent performance regardless of cost design, confirming that coordination is encoded in the demonstrations rather than induced by the cost.

\para{Key Takeaways}
This result empirically confirms the design-space analysis of \cref{sec:design-space}: the cost function must compensate for the coordination gap between the single-agent base policies and the target multi-agent behavior---%
a gap that multi-agent imitation learning encodes in the demonstration data in a way that single-agent demonstrations cannot~\textbf{(C6)}.

\section{Conclusion}\label{sec:conclusion}

This work introduces \ac{codi}, a method for synthesizing coordinated multi-agent behavior from single-agent demonstrations.
By guiding the product of independently trained single-agent diffusion policies with a multi-agent cost function, \ac{codi} bridges the gap between uncoordinated single-agent behavior and effective joint coordination—without requiring any multi-agent demonstrations.

In simulation and hardware experiments on a two-arm manipulation task, \ac{codi} discovers coordinated behavior by combining single-agent demonstrations with a cost-based guidance signal, demonstrating greater data efficiency than multi-agent imitation learning baselines.
Our ablation studies highlight the importance of joint guidance, the impact of base-policy support, and the role of cost design in achieving reliable coordination.

In this work, we examined how far performance can be pushed under the extreme data constraint of \emph{no} multi-agent demonstrations.
In many practical scenarios, however, \emph{some} multi-agent demonstrations may be available.
This raises a natural question: how should such data be used, and how should it be combined with single-agent demonstrations?
For example, multi-agent demonstrations could be used to fine-tune a base policy pretrained on single-agent data or to learn the guidance term directly, removing the need for explicit cost specification.
Exploring these directions is a natural and compelling avenue for future work.

Furthermore, the current version of \ac{codi} relies on centralized execution to coordinate single-agent policies at test-time.
Future work should explore how to distill a decentralized multi-agent policy from our approach.

Finally, this work assumes fully \emph{cooperative} behavior, in which all agents optimize a shared cost function under correlated joint action distributions.
Extending \ac{codi} to the \emph{non-cooperative} setting---encompassing agents with distinct objectives and independent action distributions---constitutes another important avenue for future investigation.

\bibliographystyle{IEEEtranN}
\bibliography{glorified,references}

{
  \appendices
  \crefalias{section}{appendix}
  \crefalias{subsection}{subappendix}
  \section{Derivations}\label{appendix:derivations}

\subsection{Optimally Compensating Multi-Agent Cost Function}
\label{appendix:optimally-compensating-multi-agent-cost-function}

For completeness, we begin by repeating the decomposition result of~\cref{eq:kl-decomposition}, labeling the second term as $\guidancecompensation\left[\guide\right]$:
\begin{align*}
   & \kldiv{\matargetpolicy}{\madiffusionpolicy} =                          \\
   & \quad \underbrace{\expectation_{\jointaction\sim\matargetpolicy}\left[
      \log\frac{\eqnmarkbox[gray]{b6}{\matargetpolicy(\jointaction\given\jointstate)}}{
        \eqnmarkbox[intent-blue]{b6}{\madatadriven(\jointaction\given\jointstate)}}
      \right]}_{\kldiv{\matargetpolicy}{\madatadriven} \geq 0}
  +
  \underbrace{\eqnmarkbox[intent-teal]{b6}{
      \log\oursnormalizingconstant +
      \expectation_{\jointaction\sim\matargetpolicy}\left[
        \frac{\guide(\jointstate,\jointaction)}{\oursregularizationfactor}
        \right]
    }}_{\defeq\eqnmarkbox[intent-teal]{b6}{\guidancecompensation\left[\guide\right]}}.
\end{align*}

To achieve a perfect match of the target behavior, i.e., $\kldiv{\matargetpolicy}{\madiffusionpolicy}=0$, we need $\matargetpolicy(\jointaction\given\state)=\madiffusionpolicy(\jointaction\given\state)$ for \emph{all} $\jointaction\in\tilde\actionspace\subseteq\jointactionspace$ for which $\tilde\actionspace$ has non-zero measure under $\matargetpolicy$. Solving for $\guide$, this yields the optimally compensating multi-agent cost function
\begin{align}
  \matargetpolicy(\jointaction\given\state) & = \frac{1}{\oursnormalizingconstant(\state)}
  \exp\left(\frac{-\guide^*(\jointaction;\state)}{\oursregularizationfactor}\right)
  \madatadriven(\jointaction\given\state)                                                                      \\
  \implies
  \guide^*(\jointaction;\state)             & = -\oursregularizationfactor\log\oursnormalizingconstant(\state)
  +\oursregularizationfactor\log\frac{\madatadriven(\jointaction\given\state)}{\matargetpolicy\left(\jointaction\given\state\right)}
\end{align}

\subsection{Decomposition of $\madiffusionpolicy$}
\label{appendix:decomposition-of-madiffusionpolicy}
\begin{align}
   & \madiffusionpolicy_{t}(\jointaction(t);\jointstate)                            \\
   & =
  \int_{\jointactionspace}
  \madiffusionpolicy(\jointaction\given\jointstate) \perturbationkernel{\jointaction}
  \intd{\jointaction}                                                    \nonumber  \\
   & =
  \madatadriven[t](\jointaction(t)\given\jointstate)
  \frac{
  \int_{\jointactionspace}
  \madiffusionpolicy(\jointaction\given\jointstate) \perturbationkernel{\jointaction}
  \intd{\jointaction}
  }{
  \madatadriven[t](\jointaction(t)\given\jointstate)
  }                                                                      \nonumber  \\
   & =
  \scalebox{1}{$\madatadriven[t](\jointaction(t)\given\jointstate)
  \int_{\jointactionspace}
  \frac{
    \exp\left(\frac{-\guide(\jointaction;\jointstate)}{\oursregularizationfactor}\right)
    \madatadriven(\jointaction\given\jointstate)
  }{
    \oursnormalizingconstant(\jointstate)
  }
  \frac{
    \perturbationkernel{\jointaction}
  }{
    \madatadriven[t](\jointaction(t)\given\jointstate)
    }
  \intd{\jointaction}$}                                                   \nonumber \\
   & =
  \madatadriven[t](\jointaction(t)\given\jointstate)
  \int_{\jointactionspace}
  \frac{
    \exp\left(\frac{-\guide(\jointaction;\jointstate)}{\oursregularizationfactor}\right)
  }{
    \oursnormalizingconstant(\jointstate)
  }
  \madatadriven(\jointaction\given \jointaction(t), \jointstate, t)
  \intd{\jointaction}                                                     \nonumber \\
   & =
  \madatadriven[t](\jointaction(t)\given\jointstate)
  \frac{
    1
  }{
    \oursnormalizingconstant(\jointstate)
  }
  \expectation_{\jointaction\sim \madatadriven(\jointaction\given \jointaction(t), \jointstate, t)}\left[
    \exp\left(\frac{-\guide(\jointaction;\jointstate)}{\oursregularizationfactor}\right)
    \right],\nonumber
\end{align}
and thus we can express the multi-agent score as
\begin{align}
  \label{eq:multi-agent-score-appendix}
   & \nabla_{\jointaction(t)} \log \madiffusionpolicy_{t}(\jointaction(t);\jointstate) =       \\
   & ~\qquad\eqnmarkbox[intent-blue]{b7}{
  \nabla_{\jointaction(t)} \log \madatadriven[t](\jointaction(t)\given\jointstate)
  }
  - \cancel{\nabla_{\jointaction(t)} \log \oursnormalizingconstant(\jointstate)}\nonumber \\
   & \quad
  +
  \underbrace{\eqnmarkbox[intent-teal]{b8}{
      \nabla_{\jointaction(t)} \log \expectation_{\jointaction\sim \madatadriven(\jointaction\given \jointaction(t), \jointstate, t)}\left[
        \exp\left(\frac{-\guide(\jointaction;\jointstate)}{\oursregularizationfactor}\right)
        \right]
    }}_{\guidancescore(\jointaction(t);\jointstate,t)\defeq}.\nonumber
\end{align}

\subsection{Simplification of $\guidancescore(\jointaction(t);\jointstate,t)$}
\label{appendix:simplification-of-guidancescore}

We can simplify
\begin{align}
   & \guidancescore(\jointaction(t);\jointstate,t)                            \\
   & =
  \nabla_{\jointaction(t)} \log
  \expectation_{\jointaction\sim \madatadriven(\jointaction\given \jointaction(t), \jointstate, t)}\left[
    \exp\left(\frac{-\guide(\jointaction;\jointstate)}{\oursregularizationfactor}\right)
  \right]                                                           \nonumber \\
   & =
  \frac{
    \nabla_{\jointaction(t)}
    \expectation_{\jointaction\sim \madatadriven(\jointaction\given \jointaction(t), \jointstate, t)}\left[
      \exp\left(\frac{-\guide(\jointaction;\jointstate)}{\oursregularizationfactor}\right)
      \right]
  }{
    \expectation_{\jointaction\sim \madatadriven(\jointaction\given \jointaction(t), \jointstate, t)}\left[
      \exp\left(\frac{-\guide(\jointaction;\jointstate)}{\oursregularizationfactor}\right)
      \right]
  }                                                                 \nonumber \\
   & =
  \scalebox{0.9}{$\displaystyle
      \frac{
        \expectation_{\jointaction\sim \madatadriven(\jointaction\given \jointaction(t), \jointstate, t)}\left[
          \exp\left(\frac{-\guide(\jointaction;\jointstate)}{\oursregularizationfactor}\right)
          \nabla_{\jointaction(t)}
          \log \madatadriven(\jointaction\given \jointaction(t), \jointstate, t)
          \right]
      }{
        \expectation_{\jointaction\sim \madatadriven(\jointaction\given \jointaction(t), \jointstate, t)}\left[
          \exp\left(\frac{-\guide(\jointaction;\jointstate)}{\oursregularizationfactor}\right)
          \right]
      }$}\nonumber
\end{align}
Due to the identities
\begin{align}
   & \nabla_{\jointaction(t)} \log \madatadriven(\jointaction\given \jointaction(t), \jointstate, t) =                                                                   \\
   & \qquad\nabla_{\jointaction(t)} \log \perturbationkernel{\jointaction} - \nabla_{\jointaction(t)} \log \madatadriven[t](\jointaction(t)\given \jointstate)\nonumber
\end{align}
and
\begin{align}
   & \nabla_{\jointaction(t)} \log \madatadriven[t](\jointaction(t)\given \jointstate) =                          \\
   & \qquad\expectation_{\jointaction\sim \madatadriven(\jointaction\given \jointaction(t), \jointstate, t)}\left[
    \nabla_{\jointaction(t)} \log \perturbationkernel{\jointaction}
    \right]\nonumber
\end{align}
we conclude that
\begin{align}
   & \guidancescore(\jointaction(t);\jointstate,t) =                                                              \\
   & \quad\expectation_{\jointaction\sim \madatadriven(\jointaction\given \jointaction(t), \jointstate, t)}\left[
    \weight(\jointaction(t), \jointaction, \jointstate, t) \nabla_{\jointaction(t)} \log \perturbationkernel{\jointaction}
    \right]\nonumber
\end{align}
where
\begin{align}
  \weight(\jointaction(t), \jointaction, \jointstate, t) \defeq
  \frac{
    \exp\left(\frac{-\guide(\jointaction;\jointstate)}{\oursregularizationfactor}\right)
  }{
    \expectation_{\jointaction\sim \madatadriven(\jointaction\given \jointaction(t), \jointstate, t)}\left[
      \exp\left(\frac{-\guide(\jointaction;\jointstate)}{\oursregularizationfactor}\right)
      \right]
  }-1
\end{align}

\subsection{Gaussian approximation of the posterior $\madatadriven(\jointaction\given \jointaction(t), \jointstate, t)$}
\label{appendix:gaussian-approximation-of-posterior}
We can obtain a Gaussian approximation of the posterior $\madatadriven(\jointaction\given \jointaction(t), \jointstate, t)$ by exploiting Tweedie's formula~\cite{robbins1992empirical,efron2011tweedie} to recover the first two moments.
\begin{align*}
  \expectation_{\jointaction\sim \madatadriven(\jointaction\given\jointaction(t),\jointstate, t)}\left[
    \jointaction
  \right] & = \jointaction(t) + t^2 \nabla_{\jointaction(t)} \log \madatadriven[t](\jointaction(t)\given \jointstate) \\
  \covariance_{\jointaction\sim \madatadriven(\jointaction\given \jointaction(t), \jointstate, t)}\left[\jointaction\right]
          & = t^2 \identity + t^4 \nabla_{\jointaction(t)}^2 \log \madatadriven[t](\jointaction(t)\given \jointstate)
\end{align*}
Recall that the score model satisfies $\mascoremodel(\jointaction(t); t, \jointstate) \approx \nabla_{\jointaction(t)} \log \madatadriven[t](\jointaction(t)\given \jointstate)$.
Hence, we can approximate
\begin{align}
  \madatadriven(\jointaction\given \jointaction(t), \jointstate, t) & \approx
  \normal(
  \jointaction;
  \muonlineguidance,
  \sigmaonlineguidance
  ),                                                                                                                                                                               \\
  \text{where }\muonlineguidance                                    & \defeq \jointaction(t) + t^2 \mascoremodel(\jointaction(t); t, \jointstate), \nonumber                       \\
  \sigmaonlineguidance                                              & \defeq t^2 \identity + t^4 \nabla_{\jointaction(t)} \mascoremodel(\jointaction(t); t, \jointstate) \nonumber
\end{align}
The Jacobian of the score model, $\nabla_{\jointaction(t)} \mascoremodel(\jointaction(t); t, \jointstate)$, is computed using automatic differentiation.

The formula for the first moment arises by expanding
\begin{subequations}
  \begin{align}
     & \nabla_{\jointaction(t)} \log \madatadriven[t](\jointaction(t)\given \jointstate) \\
     & =
    \frac{
    \nabla_{\jointaction(t)} \madatadriven[t](\jointaction(t)\given \jointstate)
    }{
    \madatadriven[t](\jointaction(t)\given \jointstate)
    }                                                                                     \\
     & =
    \frac{
    \int_{\jointactionspace} \madatadriven(\jointaction\given\jointstate) \nabla_{\jointaction(t)} \overbrace{\normal(\jointaction(t)\given \jointaction, t^2\identity)}^{\perturbationkernel{\jointaction}} \intd\jointaction
    }{
    \madatadriven[t](\jointaction(t)\given \jointstate)
    }                                                                                     \\
     & =
    \frac{
    \int_{\jointactionspace} \madatadriven(\jointaction\given\jointstate) (\jointaction - \jointaction(t))\normal(\jointaction(t)\given \jointaction, t^2\identity) \intd\jointaction
    }{
    t^2 \madatadriven[t](\jointaction(t)\given \jointstate)
    }                                                                                     \\
     & =
    \frac{
      1
    }{
      t^2
    }
    \expectation_{\jointaction\sim \madatadriven(\jointaction\given\jointstate)}\left[
      \jointaction - \jointaction(t)
      \right],
  \end{align}
\end{subequations}
revealing
\begin{align*}
  \expectation_{\jointaction\sim \madatadriven(\jointaction\given\jointstate)}\left[
    \jointaction
  \right] & = \jointaction(t) + t^2 \nabla_{\jointaction(t)} \log \madatadriven[t](\jointaction(t)\given \jointstate).
\end{align*}

The formula for the second moment arises by expanding
\begin{align}
   & \nabla_{\jointaction(t)}^2 \log \madatadriven[t](\jointaction(t)\given \jointstate)
  =                                                                                                                                                                                                                                                                                             \\
   & \quad\frac{1}{\madatadriven[t](\jointaction(t)\given \jointstate)} \nabla_{\jointaction(t)}^2 \madatadriven[t](\jointaction(t)\given \jointstate)\nonumber                                                                                                                               \\
   & \quad \scalebox{0.9}{$\displaystyle- \frac{1}{\madatadriven[t](\jointaction(t)\given \jointstate)^2}(\nabla_{\jointaction(t)}\madatadriven[t](\jointaction(t)\given \jointstate)) (\nabla_{\jointaction(t)} \madatadriven[t](\jointaction(t)\given \jointstate))^\transpose$}\nonumber,
\end{align}
and exploiting the identities $\frac{\nabla p}{p} = \nabla \log p$ and
\begin{align}
   & \nabla_{\jointaction(t)}^2 \madatadriven[t](\jointaction(t)\given \jointstate)                   \\
   & \qquad =
  \nabla_{\jointaction(t)} \left(
  \frac{
  \madatadriven[t](\jointaction(t)\given \jointstate)
    }{t^2}
  \expectation_{\jointaction\sim \madatadriven(\jointaction\given \jointaction(t), \jointstate, t)}\left[
      \jointaction - \jointaction(t)
      \right]
  \right)                                                                                    \nonumber \\
   & \qquad =
  \madatadriven[t](\jointaction(t)\given \jointstate)
  \left(
  -\frac{1}{t^2}I + \frac{1}{t^4} C
  \right)\nonumber
\end{align}
where
\begin{align}
  C \defeq \expectation_{\jointaction\sim \madatadriven(\jointaction\given \jointaction(t), \jointstate, t)}\left[
    (\jointaction - \jointaction(t))(\jointaction - \jointaction(t))^\transpose
    \right]
\end{align}
to recover
\begin{align}
   & \nabla_{\jointaction(t)}^2 \log \madatadriven[t](\jointaction(t)\given \jointstate) =                                                                                                                 \\
   & \quad - \frac{1}{t^2}I + \frac{1}{t^4} C                                               \nonumber                                                                                                       \\
   & \quad - (\nabla_{\jointaction(t)} \log \madatadriven[t](\jointaction(t)\given \jointstate)) (\nabla_{\jointaction(t)} \log \madatadriven[t](\jointaction(t)\given \jointstate))^\transpose.\nonumber
\end{align}
Recognizing
\begin{align}
  C       = &
  \covariance_{\jointaction\sim \madatadriven(\jointaction\given \jointaction(t), \jointstate, t)}\left[\jointaction\right]                                                                                                                                                                 \\
            & \quad + \expectation_{\jointaction\sim \madatadriven(\jointaction\given \jointaction(t), \jointstate, t)}\left[\jointaction(t)\right]\expectation_{\jointaction\sim \madatadriven(\jointaction\given \jointaction(t), \jointstate, t)}\left[\jointaction(t)\right]^\transpose
  \nonumber
\end{align}
we conclude that
\begin{align*}
  \covariance_{\jointaction\sim \madatadriven(\jointaction\given \jointaction(t), \jointstate, t)}\left[\jointaction\right]
   & = t^2 \identity + t^4 \nabla_{\jointaction(t)}^2 \log \madatadriven[t](\jointaction(t)\given \jointstate).
\end{align*}

\section{Connection to Conventional Classifier Guidance}
\label{appendix:relationship-to-conventional-classifier-guidance}
\para{Background on Classifier Guidance}
A useful property of diffusion models is that their generative process can be adapted at test-time by composing multiple score models.
A prominent instance of this composition is classifier-based guidance~\cite{dhariwal2021diffusion}, which adapts a diffusion model of an \emph{unconditional} distribution~$\ptarget(\sample)$ to instead match a new~\emph{conditional} distribution~$\classifier(\sample\given\guidecontext)$, where~$\guidecontext\in\reals^{\guidecontextdim}$ denotes test-time context information not known during training.
This is enabled by the identity
\begin{align}
  \label{eq:vanilla-classifier-guidance}
   & \nabla_{\sample(t)}\log\classifier(\sample(t)\given t,\guidecontext)                  \\
   & =
  \nabla_{\sample(t)}\log\ptarget_{t}(\sample(t))
  + \nabla_{\sample(t)}\log\classifier_{t}(\guidecontext\given\sample(t))
  \nonumber                                                                                \\
   & \quad - \cancel{\nabla_{\sample(t)}\log\classifier_{t}(\guidecontext)},\nonumber
\end{align}
where
$\classifier_{t}(\guidecontext\given\sample(t)) \defeq
  \expectation_{\sample\sim \ptarget(\sample\given\sample(t),t)}\left[
    \classifier(\guidecontext\given\sample)
    \right]$
is the noise-conditioned classifier: the likelihood of $\guidecontext$ averaged over the base model's denoising posterior at noise level $t$.
In practice, $\classifier_{t}(\guidecontext\given\sample(t))$ is estimated by regressing the clean-sample score $\classifier(\guidecontext\given\sample)$ against noisy observations $\sample(t)$ via MSE: since the MSE minimizer is the conditional expectation, this recovers $\classifier_{t}(\guidecontext\given\sample(t))$ exactly from samples of the joint $(\sample, \sample(t))$ drawn forward through the diffusion process.
Thus, given an unconditional score model $\scoremodel[\ptarget](\sample(t); t) \approx \truescore(\sample(t); t)$, we can obtain conditional samples from $\classifier(\sample\given\guidecontext)$ by adding the classifier score $\nabla_{\sample(t)}\log\classifier_{t}(\guidecontext\given\sample(t))$ to $\scoremodel[\ptarget]$ in the reverse SDE of~\cref{eq:reverse-sde-with-score-model}.

\para{Recognizing $\guidancescore$ as a classifier score}
When we interpret $\exp\left(-\guide(\jointaction;\jointstate)/\oursregularizationfactor\right)$ as a classifier $\classifier(\guidecontext\given\jointaction,\jointstate)$,
we can rewrite
{$\guidancescore(\jointaction(t);\jointstate,t)$ as the gradient of a noise-conditioned log classifier
\begin{align}
  \label{eq:guidance-score}
  \guidancescore(\jointaction(t);\jointstate,t) & \defeq
  \scalebox{0.9}{$\displaystyle
      \nabla_{\jointaction(t)} \log \expectation_{\jointaction\sim \madatadriven(\jointaction\given \jointaction(t), \jointstate, t)}\left[
        \overbrace{\exp\left(\frac{-\guide(\jointaction;\jointstate)}{\oursregularizationfactor}\right)}^{\classifier(\guidecontext\given\jointaction(t), \jointstate, t)}
  \right]$}\nonumber                                     \\
                                                & =
  \nabla_\jointaction \log \classifier(\guidecontext\given\jointaction(t), \jointstate, t)
\end{align}
making \cref{eq:multi-agent-score} reminiscent of the vanilla classifier guidance framework (cf. \cref{eq:vanilla-classifier-guidance}) with the composed score
\begin{align}
   & \nabla_{\jointaction(t)}\log \madiffusionpolicy_{t}(\jointaction(t);\jointstate)  \\
   & = \nabla_{\jointaction(t)}\log \madatadriven[t](\jointaction(t)\given\jointstate
  + \underbrace{\nabla_{\jointaction(t)}\log \classifier(\guidecontext\given\jointaction(t), \jointstate, t))}_{\guidancescore(\jointaction(t);\jointstate,t))}.\nonumber
\end{align}

\section{Cost Function Implementation Details}
\label{appendix:cost-design-details}

We use the following cost function structure, repeated here for convenience:
\begin{align}
  \label{appendix:eq:franka-cost-function}
  \scalebox{0.85}{$\displaystyle
      \guide(\jointstate,\jointaction) = \weightedcostcomponent[goal](\jointstate,\jointaction) + \weightedcostcomponent[collision](\jointstate,\jointaction) + \weightedcostcomponent[engage](\jointstate,\jointaction).
    $}\nonumber
\end{align}
Below, we provide the implementation details for this cost structure.
\Cref{tab:cost-parameters} lists the values we use for any parameters referenced below.

\para{Notation}
Let~$\jointtrajectory: \{1, \ldots, \horizon\} \to \jointstatespace$ denote the joint state trajectory resulting from applying the action-sequence~$\jointaction$ from state~$\jointstate$ over the receding-horizon window of length~$\horizon$ so that $\trajectory(k) = \jointstate_k$ denotes the joint state at future time step $k$.
Let $p_k^{(i)}\in\reals^3$ denote the position of the $i$th end-effector at time step~$k$ in the workspace and $p_k^{\text{object}}\in\reals^3$ denote the position of the object at time step~$k$.
Let $\pdist(p^\textrm{a}, p^\textrm{b})$ denote the Euclidean distance between two points $p^\textrm{a}\in\reals^3$ and $p^\textrm{b}\in\reals^3$ in the workspace.
Let $\tarjToPointDistance(\tilde\trajectory, \tilde{p})$ denote the minimum distance between any point along the trajectory $\tilde\trajectory$ and the point $\tilde{p}$, i.e.
\begin{align}
  \tarjToPointDistance(\tilde\trajectory, \tilde{p}) = \min_{k=1, \ldots, \horizon} \norm{p_k^{(i)} - \tilde{p}},
\end{align}
and analogously for a pair of trajectories
\begin{align}
  \trajToTrajDistance(\tilde\trajectory^\textrm{a}, \tilde\trajectory^\textrm{b}) = \min_{k=1, \ldots, \horizon} \norm{p_k^{(i)} - p_k^{(j)}},
\end{align}
Finally, let~$p_\text{goal}$ denote the goal position and~$p_\text{object}$ denote the position of the object associated with state~$\jointstate$.
The cost components are defined as follows.

\para{Goal Cost}
The goal cost component penalizes the minimum distance of the object to the goal along the receding-horizon trajectory
\begin{align}
  \costcomponent[goal](\jointstate,\jointaction) = \min\{\tarjToPointDistance(\jointtrajectory^{\text{object}}, p_\text{goal})-\goalradius, 0\}.
\end{align}
where $\goalradius$ denotes the threshold at which this penalty is applied.

\para{Collision-Avoidance Cost}
The collision-avoidance cost component penalizes the largest safety-distance violation between both end-effectors via
\begin{align}
  \costcomponent[collision](\jointstate,\jointaction) = \min\{\allowedminimumdistance - \trajToTrajDistance(\jointtrajectory^{(1)}, \jointtrajectory^{(2)}), 0\},
\end{align}
where $\allowedminimumdistance$ denotes the threshold at which this penalty is applied.

\para{Engagement Cost}
The engagement cost encourages the closer robot to engage with the object via
\begin{align}
   & \costcomponent[engage](\jointstate,\jointaction) \\\nonumber
   & =  \scalebox{0.8}{$
      \begin{cases}
        \min\{\trajToTrajDistance(\jointtrajectory^{(1)}, \jointtrajectory^{\text{object}}) - \engageradius, 0\} & \text{if } \text{\small$\norm{p_1^{(1)} - p_1^\text{object}} < \norm{p_1^{(2)} - p_1^\text{object}}$} \\
        \min\{\trajToTrajDistance(\jointtrajectory^{(2)}, \jointtrajectory^{\text{object}}) - \engageradius, 0\} & \text{otherwise}.
      \end{cases}
    $}
\end{align}

\begin{table}[ht]
  \centering
  \begin{tabular}{lll}
    \toprule
    \textbf{Parameter}              & \textbf{Symbol}           & \textbf{Value} \\
    \midrule
    Goal penalty threshold          & $\goalradius$             & \SI{1}{cm}     \\
    Goal reaching cost weight       & $w_\text{goal}$           & $1$            \\
    Engage penalty threshold        & $\engageradius$           & \SI{20}{cm}    \\
    Engage cost weight              & $w_\text{engage}$         & $10$           \\
    Collision avoidance radius      & $\allowedminimumdistance$ & \SI{30}{cm}    \\
    Collision avoidance cost weight & $w_\text{collision}$      & $10$           \\
    \bottomrule
  \end{tabular}
  \caption{Cost function parameters.}
  \label{tab:cost-parameters}
\end{table}

\section{Baselines}
\label{appendix:baselines}

Unlike \ac{codi}, the multi-agent baselines discussed below require access to a multi-agent demonstration dataset $\madataset$; both to pre-train the base policy score model $\scoremodel$ via imitation learning on $\madataset$, and to supply the state samples used in the fine-tuning losses below.
Each baseline fine-tunes this pre-trained policy by representing the score of its target policy via a residual structure:
\begin{align}
  \label{eq:baseline-score-parameterization}
  \scoremodel[][\guideparams](\jointaction(t);\jointstate,t)
  = \underbrace{\scoremodel(\jointaction(t); t, \jointstate)}_{\text{pre-trained base}}
  + \underbrace{g_\guideparams(\jointaction(t);\jointstate,t)}_{\text{learned residual}},
\end{align}
where $\scoremodel$ is a score model of the pre-trained base policy and $g_\guideparams$ is a learned guidance term with trainable parameters $\guideparams$.
While sharing this high-level structure, the baselines differ in how $g_\guideparams$ is parameterized and trained, as detailed below.

\subsection{Multi-Agent Classifier Guidance}
\label{appendix:baselines-cg}

The multi-agent classifier guidance baseline seeks to generate samples from the distribution
\begin{align}
  \madiffusionpolicy(\jointaction\given\jointstate) \propto p(\jointaction\given\jointstate)\exp\!\left(-\frac{\cost(\jointstate,\jointaction)}{\oursregularizationfactor}\right),
\end{align}
where $p(\jointaction\given\jointstate)$ is the base-policy learned from \emph{multi-agent} demonstrations.
As established in~\cref{appendix:relationship-to-conventional-classifier-guidance}, this is achieved by sampling from a diffusion model with $\scoremodel[][\guideparams]$ as in~\cref{eq:baseline-score-parameterization} and $g_\guideparams = \nabla_{\jointaction(t)}\log\classifier_t$, where $\classifier_t(\guidecontext\given\jointaction(t),\jointstate)=\expectation_{\jointaction\sim\madatadriven(\jointaction\given\jointaction(t),\jointstate,t)}\!\left[\exp(-\cost/\oursregularizationfactor)\right]$ is the noise-conditioned classifier from~\eqref{eq:guidance-score}.

We approximate the noise-conditioned classifier by learning a noise-conditioned cost model by minimizing
\begin{align}\label{eq:macg-cost-model-loss}
    \costmodelloss(\guideparams) = \expectation_{
      \substack{
        (\jointstate,\jointaction)\sim\madataset\\
        t\sim\sigmadistribution\\
        \noise\sim\normal(\mathbf{0}, \identity)
    }} \left[ \costmodel(\jointstate,\jointaction + t\noise; t) - \cost(\jointstate,\jointaction) \right]^2.
\end{align}
and use its log-gradient (obtained via automatic differentiation) to approximate the guidance score as
\begin{align}\label{eq:macg-guidance-score}
    g_\guideparams(\jointaction(t);\jointstate,t) &= \nabla_{\jointaction(t)} \log \classifier_{t}(\guidecontext\given\jointaction(t), \jointstate)\\ & \approx
    \frac{-\nabla_{\jointaction(t)} \costmodel(\jointstate,\jointaction(t); t)}{\oursregularizationfactor}.
\end{align}

\subsection{Diffusion-Policy Mirror Descent}
\label{appendix:baselines-dpmd}

\ac{dpmd}~\citep{ma2025efficient} iteratively updates a diffusion policy via mirror descent.
At each iteration, it solves the same KL-regularized problem as \cref{eq:joint-optimization-problem}, but regularizes towards the \emph{previous} policy iterate instead of the data-driven prior:
\begin{align}
  \pi^\text{MD} \in \arg\min_{\tilde\policy}
  \expectation_{\jointaction\sim\tilde\policy}\!\left[\guide(\jointstate, \jointaction)\right]
  - \oursregularizationfactor\kldiv{\tilde\policy}{\policy^\text{old}},
\end{align}
which yields the mirror descent update
\begin{align}
  \policy^\text{MD}(\jointaction \given \jointstate)
  \propto \policy^\text{old}(\jointaction \given \jointstate)
  \exp\!\left(\frac{-\guide(\jointstate, \jointaction)}{\oursregularizationfactor}\right).
\end{align}
The authors show that the score of target policy $\pi^\text{MD}$ can be learned by minimizing the cost-weighted score matching loss
\begin{align}
  &\loss_\text{DPMD}(\guideparams)
  \propto\nonumber\\
  &\quad \expectation_{\substack{
    \jointstate \sim \madataset \\
    \jointaction \sim \policy^\text{old}(\cdot\given\jointstate) \\
    t \sim \sigmadistribution \\
    \noise \sim \normal(\mathbf{0}, \identity)
  }}\!\left[
    \exp\!\left(\frac{-\guide(\jointstate, \jointaction)}{\oursregularizationfactor}\right)
    \norm{
      t^2\scoremodel[][\guideparams](\jointaction + t\noise;\jointstate,t) + t\noise
    }^2
  \right].
\end{align}

In our implementation of \ac{dpmd} and all following baselines, we parameterize the guidance term $g_\guideparams$ using the same denoising U-Net architecture as the base policy.

\subsection{Soft Diffusion Actor-Critic}
\label{appendix:baselines-sdac}
\ac{sdac}~\citep{ma2025efficient} seeks a policy that minimizes the cost while also regularizing towards maximum entropy, reducing the problem to:
\begin{align}
  \pi^\text{soft}\in\arg\min_{\tilde\policy}
  \expectation_{\jointaction\sim\tilde\policy}\!\left[\guide(\jointstate, \jointaction)\right]
  - \oursregularizationfactor H(\tilde\policy),
\end{align}
where $H(\tilde\policy) \defeq -\expectation_{\jointaction\sim\tilde\policy}\!\left[\log \tilde\policy(\jointaction\given \jointstate)\right]$ is the Shannon entropy of $\tilde\policy$.
The solution is the soft-optimal policy
\begin{align}
  \policy^\text{soft}(\jointaction \given \jointstate)
  \propto \exp\!\left(\frac{-\guide(\jointstate, \jointaction)}{\oursregularizationfactor}\right).
\end{align}
\ac{sdac} approximates the soft optimal policy by minimizing the reweighted score matching loss
\begin{align}
  & \loss_\text{SDAC}(\guideparams)
  \propto\nonumber\\
  &\expectation_{\substack{
    \jointstate \sim \madataset \\
    t \sim \sigmadistribution \\
    \jointaction(t) \sim h_t(\cdot\given\jointstate) \\
    \tilde\noise \sim \normal(\mathbf{0}, \identity)
  }}\!\left[
    \exp\!\left(\frac{-\guide(\jointstate, \jointaction(t) + t\tilde\noise)}{\oursregularizationfactor}\right)
    \norm{
      t^2\mascoremodel[\guideparams](\jointaction(t);\jointstate,t) + t\tilde\noise
    }^2
  \right],
\end{align}
here adapted for the variance schedule $\sigma(t) = t$ used by our base policy as in~\cref{sec:background}.
In this formulation, $h_t$ is a user-defined auxiliary distribution.
In our experiments we obtain samples from $h_t$ by corrupting rollouts from the current policy $\policy^\text{old}$ via the forward diffusion process, i.e., $\jointaction(t) = \jointaction + t\noise$ with $\jointaction \sim \policy^\text{old}(\cdot\given\jointstate)$ and $\noise \sim \normal(\mathbf{0}, \identity)$, as suggested in \cite{ma2025efficient}.

\subsection{Expressive Policy Optimization}
\label{appendix:baselines-expo}
\ac{expo}~\citep{dong2025expo} iteratively fine-tunes the diffusion policy by generating locally improved synthetic demonstrations and distilling them back via score matching.
At each iteration, an \emph{edit policy} $\hat\pi(\hat\jointaction \given \jointstate, \jointaction)$ proposes action-space edits $\hat\jointaction$ around each base sample $\jointaction \sim \policy^\text{old}(\cdot\given\jointstate)$, yielding improved synthetic demonstrations $\tilde\jointaction = \jointaction + \hat\jointaction$.
These are distilled into the diffusion policy by minimizing
\begin{align}
  \loss_\text{EXPO}(\guideparams)
  \propto
  \expectation_{\substack{
    \jointstate \sim \madataset \\
    \jointaction \sim \policy^\text{old}(\cdot\given\jointstate) \\
    \hat\jointaction \sim \hat\pi(\cdot\given\jointstate,\jointaction) \\
    t \sim \sigmadistribution \\
    \noise \sim \normal(\mathbf{0}, \identity)
  }}\!\left[
    \norm{
      t^2\scoremodel[][\guideparams](\tilde\jointaction + t\noise;\jointstate,t) + t\noise
    }^2
  \right].
\end{align}
In our implementation, we approximate $\hat\pi$ via \ac{mppi}: given $\nmppisamples$ perturbations $\hat\jointaction_i \sim \normal(\mathbf{0}, \Sigma)$, the edit is the importance-weighted combination
\begin{align}
  \hat\jointaction
  = \sum_{i=1}^{\nmppisamples} w_i \hat\jointaction_i,
  \qquad
  w_i \propto \exp\!\left(\frac{-\guide(\jointstate, \jointaction + \hat\jointaction_i)}{\oursregularizationfactor}\right),
\end{align}
which avoids a nested learning loop to train a separate policy network.

}

\end{document}